# OnSight Pathology: A real-time platform-agnostic computational pathology companion for histopathology

**Jinzhen Hu**[1,2*], **Kevin Faust**[2*], **Parsa Babaei Zadeh**[2*], **Adrienn Bourkas**[1], **Shane Eaton**[1], **Andrew Young**[2], **Anzar Alvi**[2], **Dimitrios George Oreopoulos**[2], **Ameesha Paliwal**[1,2], **Assem Saleh Alrumeh**[1,3], **Evelyn Rose Kamski-Hennekam**[2] & **Phedias Diamandis**[1-4]

[1]Department of Laboratory Medicine and Pathobiology, University of Toronto, Toronto, ON, M5S 1A8, Canada.
[2]Princess Margaret Cancer Centre, 101 College Street, Toronto, ON, M5G 1L7, Canada.
[3]Laboratory Medicine Program, Department of Pathology, University Health Network, 200 Elizabeth Street, Toronto, ON, M5G 2C4, Canada.
[4]Department of Medical Biophysics, University of Toronto, Toronto, ON, M5S 1A8, Canada.

* equal contribution

**Please direct correspondence to:**
Phedias Diamandis, MD, PhD, FRCPC
Neuropathologist, Department of Pathology, University Health Network
12-308, Toronto Medical Discovery Tower (TMDT), 101 College St, Toronto, M5G 1L7
p.diamandis@mail.utoronto.ca

**The microscopic examination of surgical tissue remains a cornerstone of disease classification, but relies on subjective interpretations and access to highly specialized experts, which can compromise accuracy and clinical care. While emerging breakthroughs in artificial intelligence (AI) offer promise for automated histological analysis, the growing number of proprietary digital pathology solutions has created significant barriers to real-world deployment. To address these challenges, we introduce OnSight Pathology, a platform-agnostic computer vision software that uses continuous custom screen captures to provide relevant real-time AI inferences to users as they review digital slide images. Accessible as a single, self-contained executable file (https://onsightpathology.github.io/), OnSight Pathology operates locally on consumer-grade personal computers without the need for complex software integration, thereby enabling cost-effective and secure deployment in both research and clinical workflows. Here we demonstrate the utility of OnSight Pathology using over 2,500 publicly available whole slide images across different slide viewers, as well as cases from our own clinical digital pathology setup. The software's robustness is highlighted across a number of routine histopathological tasks, including the classification of common brain tumor types, mitosis detection, and the quantification of immunohistochemical stains. A built-in multi-modal chat assistant also provides verifiable descriptions of images, free of rigid class labels, for additional quality control. Lastly, we show the compatibility of OnSight Pathology with live microscope camera feeds, including from personal smartphones, offering potential for deployment in more analog, inter-operative and telepathology settings. Together, we highlight how OnSight Pathology can deliver real-time AI inferences across a broad range of pathology pipelines, effectively removing critical barriers to the adoption of machine learning tools in histopathology.**

Despite rapid advances in non-invasive molecular profiling of patients' lesions from blood, the histopathological examination of tissue, using a microscope, has remained a cornerstone for modern disease classification[1]. For neoplasms, hematoxylin and eosin (H&E)-stained tissue slides facilitate rapid cancer subtyping and grading through detection and quantification of distinctive cytoarchitectural patterns and cellular features (e.g. mitoses)[2–4]. Analysis of H&E-stained sections also offers a dynamic tool to guide cost-effective triaging of the most appropriate molecular studies for each clinical context. For many cancer types, biomarkers that predict outcome and treatment response rely on accurate quantification of specific immunohistochemistry (IHC)-stained tissue sections[5–7]. For example, MIB1/Ki-67 staining, which provides a readout of the fraction of proliferating cells, has been linked to tumor aggressiveness[8–10]. Despite this central role, the qualitative nature of tissue examination is prone to subjective interpretative variability, especially in resource-strained centers, which can compromise precision care[11].

Over the past decade, fueled by breakthroughs in deep learning and the digitization of tissue sections into whole slide images (WSIs), many of these manual exercises have proven capable of being objectified and automated through computational analysis[12,13]. In addition to reaching expert-level performance on many traditional tasks, the application of AI and machine learning to large public and private catalogs of WSIs has allowed for the generation of high-dimensional "histomic" signatures that can predict the underlying mutational status of tumors and risk scores that correlate with divergent clinical outcomes[14,15]. Despite these exciting advances, adoption has stalled; partly due to multiple technological barriers[16–18]. Only recently has high-quality and large-scale slide scanning become available to allow the transition of entire departments into digital pathology centers capable of exploiting available AI solutions. This rapid progress has come from a variety of digital scanner manufacturers, many with their own proprietary WSI formats and slide-viewing software, which complicates integration[19,20] (**Fig 1a**). Moreover, the extreme diversity of AI algorithms adds additional nuances as to which AI models should be deployed to which subset of slide and tissue regions *a priori*. Independent self-contained local (e.g. QuPath[21]) and cloud-based (e.g. DeepLIIF[22]) solutions do exist, but these often require users to have access to WSIs and/or manually extract regions of interest (ROI) for input; which can interrupt workflows and risk a breach of sensitive personal health information (PHI) (**Fig 1b**). Platform-agnostic tools that provide compatibility across different digital setups and allow microscopists to dictate which AI algorithms are deployed to which subregions of available tissue will likely help facilitate more widespread adoption of computational pathology.

Previous work attempting to address these implementation barriers proposed an "augmented reality microscope" that uses custom camera hardware, installed on a conventional microscope, to continuously capture images from mounted glass slides[2]. These images are run through various AI models, with inferences provided to the user, in real-time (defined as a ~200-400 millisecond refresh rate)[2], by superimposing results within the microscope's eyepiece field of view (FOV). As a proof-of-concept, the authors highlighted their tool's utility in cancer detection, and envisioned additional future applications for classification, object detection and quantification of both H&E and IHC-stained sections. While this prototype highlighted how many key adoption barriers could be addressed by outfitting individual microscopes with reasonably accessible hardware, the recent focus on transitioning pathology to fully digital clinical workflows has created opportunities to further extend this concept to more pathologists and at even lower costs of entry. Towards this, we developed an open and freely downloadable software we coined "OnSight Pathology" that utilizes customized continuous screen capture, from a personal computer's display, as input, to provide real-time AI inferences as users review WSIs (**Fig 2a, Supplementary Video 1**). To highlight compatibility across diverse environments, we show how OnSight Pathology can aid during routine histopathological tasks by benchmarking it using both public cohorts and our own local real-world clinical workflow. While we use various accessible consumer-grade laptops with dedicated GPUs (~$1000USD) to produce "real-time" inference speeds across relevant tasks, the software is also compatible with standard CPU-based computers (**Fig 2b**).

With a broad base of biomedical end users in mind, we designed OnSight Pathology to require no significant coding expertise or specialized hardware for use.

OnSight Pathology is also agnostic to differences in proprietary WSI and slide viewer formats and all computer vision inference generation occurs "on site"; avoiding external transfer of sensitive patient PHI data (**Fig 2c**). To achieve this, we packaged OnSight Pathology into a one-click downloadable and executable file (.exe) using PyInstaller and Inno Setup to eliminate the need for users to manage complex dependencies or Python environments (available at https://onsightpathology.github.io/). Once installed, OnSight Pathology's graphical user interface (GUI), built with the PyQt framework, allows users to define their display's region of interest for continuous image extraction, customize inference settings, and seamlessly switch between different AI models without needing to modify or interact with any code (**Fig 2a**). The underlying multi-threaded backend pipeline leverages the pynput (v1.8.1) and mss (v10.0.0) libraries for efficient region selection and screen capture, with the OpenCV (v4.11.0) library handling continuous image handling and preprocessing before routing frames to the local model inference engine. This design makes OnSight Pathology's functionality entirely independent of any proprietary vendor software requirements and avoids a reliance on any application programming interfaces (APIs) that may require an "off-site" transfer of images for analysis. Because the displayed FOV often comprises a relatively small fraction of the overall WSI, we show even modest personal computer hardware is sufficient to output AI-based analyses in near "real-time". This "on site" analysis also helps overcome the many privacy concerns associated with using external online AI tools or APIs for sensitive patient data. For tasks that require surveying of larger and/or specific sizes of tissue regions, we include an "aggregate mode" in OnSight Pathology where AI inferences over multiple high-power fields of view can be recorded and summed together (**Supplementary Fig 1**). This includes a built-in calibration tool that takes advantage of the slide viewers' ruler to help standardize outputs from pixel densities to physical areas (e.g. mitoses per mm²) (**Supplementary Fig 2**).

      OnSight Pathology is compatible and customizable with a wide variety of existing public and custom-trained models and architectures. The initial release, associated with this description, includes classification, object detection, and segmentation models; which, in aggregate, make up a large fraction of computational pathology solutions. In addition to validating its strong performance across these supervised tasks, we demonstrate OnSight Pathology's compatibility with emerging Medical Domain Vision-Language Models (HuaTuo & LingShu)[23,24] that supports interactive, natural-language querying of specific ROI. We highlight how real-time access to these more verifiable descriptive AI inferences can help overcome the limitations of label space rigidity that limit the applicability of many supervised classification models in the field.

      OnSight Pathology is pre-loaded with models capable of classifying common brain tumor types, performing mitosis detection, and quantifying IHC stains (**Fig 3a-c**). To demonstrate real-time classification of common brain tumor histologies, we employed a Vision Transformer (ViT-B/16) backbone, initialized with weights from Kaiko.ai's domain-specific foundation model for digital pathology, trained via a self-supervised DINOv2 algorithm on The Cancer Genome Atlas Program (TCGA) WSIs[25].

The model was then fine-tuned on a dataset comprising 80,000 H&E-stained image tiles (1,024 × 1,024 pixel), captured at 20x magnification (~0.5 µm/pixel), from our own cancer center (University Health Network, Toronto). These images were distributed evenly across the four most common tumor patterns encountered in neuropathology: (i) glial, (ii) meningothelial, (iii) epithelial and (iv) schwannian histology and meant to help guide the initial classification of surgical pathology specimens. Extensive photometric augmentations were applied during training to simulate stain variability, with the aim of enhancing generalizability across a wide spectrum of clinical applications, including digitized WSIs of conventional formalin-fixed paraffin embedded (FFPE) tissue, intra-operative frozen tissue sections embedded within optimal cutting temperature (OCT) compound and live microscope imaging.

To benchmark the model's performance within OnSight Pathology setup, we first generated classification scores across 1,620 public WSIs from the EBRAINS resource[26] (**Fig 3d-e**). This cohort included 943 gliomas, 344 meningiomas, 47 epithelial metastatic carcinomas, and 75 schwannomas. An additional 211 WSIs representing tumor types outside these categories (e.g. hemangioblastomas, lymphomas) were included to assess how the model handled out-of-distribution samples. All WSIs were downloaded and viewed through the Open QuPath WSI viewer[21]. Overall, in-distribution classification accuracy reached 0.98, with multi-class area under the multi-class receiver operating characteristic (ROC) curves (AUC) being 0.97 or higher across the included tumor types. Confidence scores for out-of-distribution lesions were consistently lower than those for correctly classified in-distribution classes (AUC = 0.93), indicating that the model can effectively flag a large proportion of atypical inputs that may warrant closer clinical evaluation (**Fig 4a**).

To demonstrate generalizability across different WSI viewing platforms, we further evaluated the model on 1,001 additional cases from TCGA, spanning the glioblastoma (TCGA-GBM) and low-grade glioma (TCGA-LGG) cohorts, as well as multiple epithelial tumor types that commonly metastasize to the brain, achieving a multi-class AUC of 0.99 (**Supplementary Fig 3**). Notably, this validation was carried out entirely using the online Digital Slide Archive viewer, without requiring local access to the original WSIs for analysis (https://cancer.digitalslidearchive.org/)[27]. We also tested model robustness on routine neuropathology surgical cases at our local cancer center, reviewed through the proprietary clinical-grade slide viewer Synapse Pathology (Fujifilm). In total, we classified 98 successive retrospective cases that were discussed at our multi-disciplinary cancer tumor board and achieved an in-distribution accuracy of 0.96 and multi-class AUC of 1.00 (**Fig 3f**, **Supplementary Fig 4**). Collectively, these findings highlight the flexibility of OnSight Pathology in deploying a state-of-the-art classification model on consumer-grade hardware, without the need for specialized software or direct access to high-resolution WSIs.

Beyond classification, H&E-stained tissue remains a central tool for tumor grading. This task includes the identification and quantification of mitoses, which can be a laborious and subjective process. To highlight how such challenging tasks could be augmented by OnSight Pathology, we therefore also included an object (mitosis) detection model within it (**Fig 3b**). Specifically, this model is derived from RetinaNet[28], which integrates a Feature Pyramid Network for multi-scale feature learning and a ResNet-18 as its convolutional feature extractor[29]. We trained the model on the MIDOG++ dataset[30], utilizing bounding-box annotations of mitotic figures at a magnification level of 40x (~0.25μm per pixel), with input image patches of size 512×512 pixels. The trained model was then embedded within the real-time processing pipeline of the OnSight Pathology system. To provide users with greater control over this task, the model's confidence score of detected mitosis was then integrated into the OnSight Pathology graphical user interface (GUI) as a user-adjustable toggle. This design enables dynamic customization of the detection threshold, allowing users to filter and prioritize higher-confidence predictions. Clinicians seeking more conservative predictions of mitoses can increase the threshold to focus on the most reliable predictions. Users can also lower it to explore additional mitotic candidates in real-time, depending on the clinical context and their personal preferences (**Supplementary Fig 5**).

To benchmark this tool, we took advantage of publicly available annotations from the International Conference on Pattern Recognition (ICPR) 2012 and 2014 Mitosis Detection Grand Challenge[31,32] to evaluate its performance. Encouragingly, we found that higher confidence scores correlated with true mitoses, yielding an AUC of 0.787 (**Fig 3g**). Using a cutoff of 0.31, OnSight Pathology yielded an F1-score (0.7539), recall (0.7422) and precision (0.7659) that was non-inferior or better than other reported tools (**Fig 3h**). Additionally, during benchmarking, we found multiple examples of OnSight detecting "false negative" mitoses, in which true mitoses were not annotated/detected in the original dataset. This underscores the potential value of OnSight Pathology for augmenting histological analysis (**Supplementary Fig 6**). Due to the importance of mitoses in grading gliomas and meningiomas, we also took advantage of our local high-quality annotations to benchmark this tool within a clinical digital pathology workflow. Using over 200 detected objects, we again found higher confidence scores were associated with true mitoses, with an AUC of 0.978 (**Fig 3i**). Overall, these results highlight the potential of OnSight Pathology to function as a copilot to improve the speed and sensitivity of mitosis detection by experts.

Quantification of immunohistochemical biomarkers, such as the Ki-67 proliferation index, represents another routine downstream clinical task that is subject to well-recognized inter-observer variability. To demonstrate the utility of OnSight Pathology's in objectifying such cell-level segmentation tasks, we therefore also trained and included a YOLO-based model[33,34] to identify and differentiate Ki-67 positive and negative nuclei (**Fig 3c**). Specifically, image patches of size 1024x1024 pixels at 20x magnification (~0.5μm per pixel) were manually selected and accompanying annotations were generated using QuPath as a ground truth (**Supplementary Fig 7**). To enhance model robustness over this existing tool, which was often challenged by images exhibiting poor contrast due to faint nuclear hematoxylin staining, targeted data augmentation techniques were applied. These included gamma correction, Gaussian blur, and controlled hue shifts within the purple-blue-cyan spectrum. Well-annotated image patches were specifically chosen to have the augmentation applied to, with the corresponding ground truth labels left unchanged. This approach showed improved detection over QuPath for challenging cases (**Supplementary Fig 8**). A transparency control that facilitates clearer visualization of overlaid segmentation masks on the underlying nuclei was also included to improve real-time interpretability during use (**Supplementary Fig 9**). To benchmark performance, we compared Ki-67 estimates of 17 public images, viewed with another online slide viewer (PathCore, Toronto), to expert estimates. Overall, this showed a high correlation (Pearson's r = 0.981) and scores similar to other readily-available tools (e.g. QuPath and DeepLIIF[22]) (**Fig 3j**). Importantly, unlike the standard workflows of these alternative tools, which require images to be extracted, saved locally and then loaded into platforms to compute scores, OnSight Pathology produced results in real-time without additional steps or workflow disruptions (**Fig 1b**). This led to substantially shorter "real-world" inference times, which we believe will help encourage its use for the routine objective quantification of Ki-67 in busy practices (**Fig 3k**). This model also performed well within our own clinical-grade workflow, showing a high correlation coefficient (Pearson's r = 0.99) with pathologist-reported scores (**Fig 3l**).

Despite their strong performance across different datasets and viewers, supervised classifiers are known to suffer from notable misclassifications when out-of-distribution classes are encountered[35]. While we found many of these errors could be flagged in OnSight Pathology due to lower confidence scores, we also observed that the high class space rigidity of the supervised ViT classification model led to some cases of diffuse large b cell lymphoma (DLBCL) and cerebral cortex being misclassified as glial histology at high confidence (**Fig 4a-g**). To show how these limitations could be addressed, we integrated a chat-based interface with multiple vision language models (VLMs) to complement OnSight Pathology's real-time visual inference pipeline. These models include the HuatuoGPT-7B Vision model, built with the PubMedVision dataset and optimized for medical and pathology applications, as well as LingShu-7B, a state-of-the-art multimodal large language model designed for medical visual question answering and clinical report generation. We hypothesized that the broad training and interpretable nature of the language-based outputs of these multi-modal models could be used as a complementary mechanism to flag potential out-of-distribution cases that otherwise produced high confidence classification scores. To test this, we asked OnSight Pathology to "provide a microscopic description of the image" for representative lymphoma, cortex and glioma cases where the ViT model scored as high-confidence glial histology (0.99). Indeed, the generated descriptions were vastly different between these classes of representative cases, with good intra-class reproducibility (**Fig 4h-j**). This highlights how users can leverage the ability to switch between multiple complementary models within OnSight Pathology to disentangle the histology of both common and rare disease types that may be encountered in less controlled real-world environments.

Lastly, while all the above validation centered around digital images extracted from WSIs, the OnSight Pathology workflow is theoretically also compatible with images from live recordings of glass slides that have been generated using microscope-mounted cameras. To highlight this generalizability and potential for telepathology applications, we therefore also tested the workflow on our department's remote digital pathology system where we report time-sensitive intra-operative frozen section diagnoses between different hospital sites (**Fig 5a**). To validate its performance, we analyzed intra-operative frozen sections from 9 prospective UHN patients, containing cases with glial (n=2), meningothelial (n=2), epithelial (n=3), and schwannian (n=2) histology. A total of 27 ROIs were analyzed from these cases, yielding an overall classification accuracy of 0.9259. OnSight Pathology notably generalized well in this workflow providing high-quality real-time interpretations without any need for further customization (**Fig 5b-d**). For users who may not have access to this hardware setup, we also demonstrate the compatibility of OnSight Pathology H&E-based models with live camera feeds captured using a smartphone mounted to a microscope via an accessible 3D-printed adapter (OpenOcular) (**Fig 5e**).

AI has the potential to provide decision support and improve objectivity and efficiency for practicing pathologists and researchers. Despite a growing number of viable applications, adoption has been met with significant barriers for routine implementation. Microscope-centered solutions require upfront investment in additional hardware, whereas modern digital pathology workflows are challenged by different proprietary file formats, storage solutions and slide viewing software; each requiring custom implementation pipelines[20]. Online platforms and API-based solutions help alleviate some of these technical challenges, but present other privacy concerns for sensitive data and poorly-suited disruptions in high volume clinical practices[36,37]. Here, we present OnSight Pathology as a real-time computational pathology companion for histopathology. OnSight Pathology can be installed and run on consumer-grade laptops with virtually no coding expertise required. Furthermore, OnSight Pathology is entirely agnostic to specific WSI file formats and viewing software, as users do not need to have direct access to WSIs to run custom analyses. Indeed, OnSight Pathology's compatibility with glass slides viewed under high-quality microscope cameras, and even mounted smartphones, extend its generalizability to additional users and clinical scenarios, including time-sensitive intra-operative consultations and resource-strained centers. Using multiple public datasets as well as our own local clinical workflow, we show how OnSight Pathology can perform relevant classification, object detection and quantification tasks in real-time, without introducing any meaningful latencies to existing processes. Moreover, results can be aggregated over multiple calibrated regions, providing outputs in highly relevant formats for immediate translatability. The wide diversity of models tested support that this platform can be adapted for additional clinical tasks and subspecialties such as tumor-specific grading and risk prediction[14,15]. Quick and simultaneous access to both traditional supervised and weakly self-supervised chat-based vision language models also offer complementary benefits to overcome the existing class space rigidity limitations of former models.

While we highlight the extremely versatile nature of OnSight Pathology, we acknowledge that there are some important caveats to consider. Despite many of the core applications tested being carried out on fairly modest laptops with near real-time performance, other larger multi-modal and language models require more powerful hardware (e.g. minimum 16 GB of GPU-based RAM) with longer compute times, which may currently limit advanced features for some users. Furthermore, for applications where WSIs need to be reviewed in their entirety, such as detection of micro-metastases in multiple lymph nodes, more automated approaches to tissue coverage have clear benefits. For many day-to-day applications however, we believe the ability of an expert to choose which AI tool(s) should be deployed to a smaller subset of slides and respective tissue regions may significantly reduce computational demands and provide additional important implementation efficiencies and advantages. Finally, formal initiatives to further validate and provide guidance as to how this tool and the included models are used in medical decision-making are required before clinical use. This is especially paramount for microscope camera-based adaptations of OnSight Pathology, where image parameters (e.g. focus, brightness) can be continuously changing during use. Overall, we anticipate the growing adoption of digital pathology and the lightweight nature of OnSlight Pathology will facilitate a more rapid and generalized adoption of AI in pathology centers.

## Methods

**Ethical Approval.** The University Health Network Research Ethics Board has approved the study REB #17-5387 as it complies with relevant research ethics guidelines and the Ontario Personal Health Information Protection Act (PHIPA, 2004). Patient consent was not directly obtained and a consent waiver for this study was granted by the University Health Network Research Ethics Board as the research was deemed to involve no more than minimal risk and included the use of exclusively unidentifiable images of existing pathological specimens. Images used to train the ViT model were generated locally and made publicly available in a previous publication[38]. The mitosis object detection model was trained using the public MIDOG++ dataset[30]. Testing on local clinical cases was done in real-time and meant to supplement the testing of larger public datasets. As such, we did not seek additional REB approval to generate new local image datasets. For the majority of benchmarking exercises, WSIs or extracted image patches were obtained from sources, including EBRAINS, TCGA and the ICPR 2012 and 2014 Mitosis Detection Grand Challenges[26,31,32,39] as well as previous publications[38] that are publicly available for research purposes. Therefore, no additional institutional review board approval was required for their use.

**Experimental Design.** The objective of this study was to develop, validate, and demonstrate the utility of OnSight Pathology, a novel, platform-agnostic software for real-time computational pathology. The study design involved: (1) The development and packaging of a standalone, multi-threaded Python application deployable as a one-click installer for both GPU and CPU-based systems. (2) The integration and optimization of deep learning models for key histopathological tasks, including tumor classification (ViT), mitosis detection (RetinaNet), and Ki-67 quantification (YOLO). (3) The incorporation of a multimodal chat assistant (VLM) for interactive analysis. (4) A multi-faceted validation strategy, benchmarking model performance against established public datasets (e.g., TCGA, EBRAINS, ICPR challenges) using various slide viewers. (5) Validating the software's performance, usability features (e.g., Aggregate Mode, Calibration), and generalizability on local, real-world clinical cohorts from the University Health Network and on live microscope camera feeds. The primary outcomes were model accuracy, concordance with expert analysis, workflow efficiency, and real-time inference performance on consumer-grade hardware.

**OnSight Pathology Architecture Design and Distribution.** OnSight Pathology is a standalone desktop application developed in Python (v3.11.9)[40] for the Windows operating system. A core design principle was to create a self-contained tool that eliminates the need for any end-user dependency management or environment configuration. To achieve this, the application and its entire software stack, including the PyTorch deep learning framework[41] and the required CUDA dynamic-link libraries, are bundled using PyInstaller (v6.13.0, https://pyinstaller.org/en/stable/) and distributed via a user-friendly installer that was created with Inno Setup (v6.5.4, https://jrsoftware.org/isinfo.php).

*Hardware Variants and Compatibility.* To support a wide range of hardware configurations, two installer variants are provided. The GPU-enabled build includes the full CUDA runtime and employs Inno Setup's disk spanning feature (split into 2 GB slices) for simplified distribution. By contrast, the CPU-only build is a fraction of the size and designed for users without compatible NVIDIA hardware. This version avoids packaging GPU-based PyTorch dynamic libraries, thereby preventing Windows from searching for non-existent CUDA DLLs that could otherwise trigger initialization errors. Both variants therefore remain as one-click executable installers, requiring no additional setup beyond installation. While the software is designed to operate on commercially available personal computers and can operate entirely on CPU-based laptops, a CUDA-capable NVIDIA GPU (e.g., GeForce RTX 10-series or later) is strongly recommended for smooth, real-time performance. To maintain compatibility with both current and future GPUs, the application is built against a PyTorch release supporting CUDA 12.8, ensuring broad support across a wide range of NVIDIA GPUs.

*GUI and Asynchronous Multi-threaded Design*. OnSight Pathology features a GUI built with the PyQt6 (v6.8.1, https://pypi.org/project/PyQt6/) framework. To ensure low latency and a responsive user experience during continuous inference, the software employs an asynchronous, multi-threaded design, dividing responsibilities between two primary threads: a main GUI thread and a background worker thread.

*Main GUI Thread and User Interface.* The main GUI thread is responsible for managing all user interactions and event handling. The interface is organized into several key functional panels, including configuration areas containing dropdown menus for selecting the AI model and the chat assistant model, along with a screen capture tool button that allows users to define the ROI. Each AI model can also expose additional customizable parameters (e.g., specificity threshold for mitosis detection), enabling users to fine-tune inference behavior without code modification. The central control panel contains the primary action buttons (Start, Stop, Export, and Chat with LLM), while the main inference display window presents the real-time annotated image feed from the model. For more advanced workflows, the interface also incorporates an "Aggregate Function" panel used for multi-FOV quantitative analysis. To enhance usability, OnSight Pathology includes a "Compact Mode", designed for scenarios where screen space is limited or when users prefer a streamlined workspace. This alternate layout vertically stacks functional panels and allows users to hide or reveal specific sections, preserving workspace visibility while retaining all essential controls.

*Background worker and thread communication.* The main GUI is intentionally kept lightweight, processing events immediately and delegating all computationally demanding operations to the background worker thread, which manages the entire data processing and inference pipeline. This architectural decoupling guarantees that computationally intensive operations never block the GUI's event loop, thereby preventing the interface from freezing and allowing for seamless user interaction while the model is running. Communication between these two threads is managed asynchronously via PyQt6's signal and slot mechanism. As the background worker thread completes each inference cycle, it emits signals containing the processed output from the AI inference, including the final annotated image and quantitative metrics, to the main thread to update the display.

*Real-time Inference Pipeline.* The process from initial screen capture to the final display of AI-generated results is structured as a continuous and sequential pipeline operating within the background worker thread. This pipeline is designed to be agnostic to the source of the digital pathology image, as it directly interacts with the pixels rendered on the user's display.

The pipeline begins with the user defining a rectangular ROI on their screen. This interaction is implemented using the pynput library (v1.8.1, https://pypi.org/project/pynput/). The library records the coordinates of two consecutive mouse clicks, corresponding to the top-left and bottom-right corners of the capture area.

Frame acquisition is then handled by the mss library (v10.0.0, https://pypi.org/project/mss/), which captures the specified ROI as a continuous stream of frames. Each captured frame, provided in BGRA format, is immediately passed to a preprocessing stage handled by the OpenCV library (v4.11.0.86, https://opencv.org/). This stage involves a color channel conversion to RGB that is required by the classification and detection models and a conversion to BGR that is required for the segmentation model. Subsequently, each frame is partitioned into inference-ready image tiles, according to OnSight Pathology's Dynamic Tiling Strategy (described below), which serve as the direct input for the deep learning models. Following inference, the generated predictions (e.g., bounding boxes, segmentation masks) are rendered as visual overlays onto the original frame. The annotated composite image is then transmitted back to the main GUI thread for display, completing the real-time feedback loop.

*Dynamic Tiling Strategy.* To accommodate user-defined ROIs of arbitrary dimensions, the pipeline incorporates a preprocessing and dynamic tiling mechanism that adapts to both undersized and oversized capture regions. If a user captures a ROI that is smaller than the model's minimum input requirement (e.g., 300x300 pixels for a model trained on 512x512 images), the frame is automatically upscaled at capture time to satisfy the minimum dimension constraints. For larger ROIs, OnSight Pathology performs patch-based inference by dividing the image into tiles whose size and overlap depend on the specific inference task. For the classification model, a non-overlapping grid is generated. To ensure analysis of the entire ROI, if the frame dimensions are not exact multiples of the tile size, the final row and column of tiles are shifted to align with the bottom and right edges of the frame. This guarantees that every pixel of the original ROI is included in the inference. For the segmentation model, a strict non-overlapping grid is used, where any residual pixels along the bottom or right edges are excluded rather than padded, ensuring every processed tile corresponds to a fully valid region of the original image and prevents edge artifacts. For mitosis detection, however, a non-overlapping grid risks missing targets that lie on the boundary between two tiles. To mitigate this, an overlapping sliding window is used.

*Tumor Classification Pipeline (VIT).* The tumor classification task is performed using a pathology foundation ViT model loaded from the timm library (v0.6.13)[42]. The model was fine-tuned on our local dataset from the University Health Network comprising 80,000 H&E-stained image tiles and operates on 1024x1024 pixel tiles.[25] Following our Dynamic Tiling Strategy, an incoming ROI is processed into a set of $N$ tiles. A small ROI that does not require partitioning is treated as a single tile ($N = 1$), while a large ROI (>1024x1024 pixels) is partitioned into multiple tiles ($N > 1$). Each tile is independently classified, producing a softmax probability vector $p_i$ over the predefined classes. A single prediction for the entire ROI is derived by computing the mean probability vector across $N$ tiles:

$$\bar{p} = \frac{1}{N} \sum_{i=1}^{N} p_i$$

This formulation handles both scenarios. For small, single-tile ROIs where $N = 1$, the equation simplifies to the probability vector of that single tile. For a large, multi-tile ROI, it computes the mean probability vector. The final predicted class $\hat{y}$ is then determined as:

$$\hat{y} = \arg\max(\bar{p})$$

This mean pooling approach aggregates tile-level confidence from the entire region, resulting in a stable and consistent global classification for the ROI.

*Mitotic Figure Detection Pipeline (RetinaNet).* Mitosis detection was implemented through a RetinaNet-based object detection model. The pipeline first processes the ROI according to our Dynamic Tiling Strategy, using an overlapping sliding-window approach. Small ROIs are processed directly as a single analytical unit, whereas larger ROIs (>512x512 pixel) are subdivided into overlapping tiles. The RetinaNet model independently analyzes each tile to identify candidate mitotic figures. For each tile, the model generates a set of candidate detections, where each detection $D_i$ is a tuple $(B_i, c_i, s_i)$ containing the bounding box coordinates $B_i$, the predicted class label $c_i$ (1 for mitotic figure, 0 for background), and an associated confidence score $s_i$. After processing all tiles, the collected set of detections are merged and refined with a custom, distance-based Non-Maximum Suppression algorithm (NMS) to eliminate duplicate detections. In this method, for each detected bounding box $B_i$, the centroid $(x_i, y_i)$ is computed, and detections are sorted by descending confidence. With a KD-tree for neighborhood search, all detections $D_j$ within a fixed radius r = 25 pixels of a higher-confidence detection $D_i$ are suppressed:

$$D_j \text{ is suppressed if } \exists D_i \text{ such that } s_i > s_j \text{ and } \sqrt{(x_i - x_j)^2 + (y_i - y_j)^2} < r$$

The variant of NMS is suited for small and cellular structures like mitotic figures, where bounding boxes may occur in close proximity without substantial overlap. The final output visualizes these confirmed detections as bounding boxes on the original image, color-coded by their confidence score: green for high confidence ($s > 0.7$), orange for medium confidence ($0.4 < s \leq 0.7$), and blue for low confidence ($s \leq 0.4$) detections.

*Ki-67 Proliferation Indexing Pipeline (YOLO).* Quantification of the Ki-67 proliferation index is achieved using a YOLO-based instance segmentation model, operating on non-overlapping 1024x1024 tiles. Following inference generation, the individual segmentation masks are stitched together to reconstruct a full-resolution segmentation map covering the entire ROI. The Ki-67 index is then calculated directly from this map as the ratio of Ki-67-positive detected cells to the total number of detected nuclei. Due to the non-overlapping tiling scheme, cells lying near the tile borders may occasionally be double-counted when partial cells extend across boundaries. However, since each tile represents a comparatively large region (1024x1024 pixels) relative to a standard computer display, the total number of tiles per ROI remains limited, minimizing this cumulative double-counting effect.

*Model Download and Cache.* OnSight Pathology employs the Hugging Face Hub library[43] to establish a secure connection to the online model repository and download the required model weights. Once downloaded, these weights are automatically stored in a local cache directory (default path: ~/.cache/huggingface/hub). On subsequent runs, OnSight Pathology loads the weights directly from this cache without re-downloading, ensuring rapid startup and offline functionality. All model inference is performed entirely locally on the user's computer.

*Crash Logging and Error Handling.* To support reliable operation across diverse hardware configurations, OnSight Pathology includes an integrated crash logging system that records and documents any runtime errors encountered during use. When an error occurs, the application halts the ongoing process, notifies the user that an error has been encountered, and specifies the location of the generated log file. These detailed logs are automatically saved to ~/AppData/Local/OnSightPathology/Logs. At present, users can manually share these log files with our team via the corresponding author contact for further analysis and debugging in cases of persistent errors.

**Aggregate Mode for Multi-FOV Quantification.** To support quantitative analysis over tissue areas that extend beyond a single FOV, we incorporated an aggregate mode that allows users to accumulate inference metrics from multiple, non-contiguous ROIs into a single analysis.

*Workflow and User-in-the-Loop Validation.* The aggregation workflow is initiated by the user through the GUI. After activating Aggregate Mode, the user can freely navigate the WSI to any desired FOV. Pressing the "spacebar" hotkey captures the AI inference metrics from the current ROI and triggers a task-specific validation step. For classification and Ki-67 quantification tasks, the validation consists of a simple accept or reject confirmation. A dialog appears that displays the model's findings, such as the predicted tumor class (with its confidence score), or the number of positive and negative cells (along with the computed Ki-67 index). This allows the user to visually inspect the ROI and the corresponding model output. Only ROIs that the user confirms as accurate are added to the final aggregation pool. For the mitosis detection task, which requires precise counting, OnSight Pathology provides a more interactive human-in-the-loop correction workflow. Pressing the spacebar hotkey presents a dialog that not only shows the model's detected mitotic figure count, but also includes an editable input field. This allows the user to directly adjust or override the count, based on expert review, before the value is committed to the aggregate total. This workflow blends automated detection with expert oversight, reducing manual effort while maintaining clinical accuracy. Upon user confirmation, the quantitative metrics are added to a running total. Each submission appends a structured entry to an in-memory session log containing the original image frame, the AI-annotated frame, and the task-specific inference metrics. The GUI provides real-time feedback throughout this process, continuously updating the cumulative statistics, including the total number of tiles analyzed, the total physical area, and derived metrics like mitotic density or the overall Ki-67 proliferation index (**Supplementary Fig 1**).

*Data Export for Post Analysis.* At the conclusion of the aggregation session, users can export all recorded results through the Export function. An output folder is generated that contains both the original image and the annotated image for every validated ROI, saved sequentially. Additionally, it generates a csv file that tabulates the metrics for each ROI and the corresponding aggregate totals. For example, for the classification tasks, the csv file includes the full softmax probability distribution across all classes for each tile (**Supplementary Fig 1**).

*Physical Unit Calibration.* To report quantitative results such as mitotic density (mitoses/mm²) in physical units rather than arbitrary pixel counts, the aggregate function includes a user-driven calibration tool for determining the physical area of each ROI. Calibration proceeds in two stages. In the first stage, OnSight Pathology overlays a fixed-dimension reference box onto the center of the live inference feed. This box occupies one-ninth of the total ROI area, with its height and width equal to one-third of the ROI's dimensions. The user then uses the measurement tools within their third-party slide-viewer software to draw a rectangle that matches the on-screen reference box, thereby obtaining its precise physical area in square millimetres. In the second stage, the user inputs this measured area into a dialog box. OnSight Pathology stores this value and establishes it as the ground-truth area for the reference box. Since the total ROI area is exactly nine times the area of this reference box, OnSight Pathology can automatically compute the physical area of any subsequent full-ROI capture (**Supplementary Fig 2**). All downstream quantitative metrics, such as mitotic density or Ki-67 index, are then reported in standard physical units. The calibration function was further validated against ground-truth measurements obtained from QuPath, with an overall Pearson correlation of 0.998 across various magnifications (**Supplementary Fig 10**).

**Interactive Multimodal Chat Assistant.** OnSight Pathology incorporates an integrated multimodal chat assistant that provides an interactive analysis of the captured ROI, extending beyond fixed classification labels. To accommodate the substantial computational and memory demands of modern VLMs, while maintaining smooth operation of the primary real-time inference pipeline, the chat module is implemented using a decoupled, out-of-process architecture.

*Architectural Design and Communication.* When the user activates the chat assistant, OnSight Pathology opens an interactive chat window as a pop-up within the main interface. The active inference thread for classification, detection, or segmentation is stopped, and its GPU memory released, allowing OnSight Pathology to allocate these resources to the VLM. User interaction with the rest of the interface is also temporarily suspended while the chat window remains open. The chat window then starts a separate worker subprocess that is responsible for loading and executing the selected VLM. This design separates the model's computational workload, which involves high VRAM and CPU usage, from the user interface layer. As a result, the chat interface itself remains responsive while the VLM generates responses, and users can safely close it at any time without causing the application to freeze or crash.

Communication between the chat window and the worker subprocess is handled asynchronously via inter-process communication. All messages, including user prompts and model-generated responses, are transmitted in lightweight JSON format to ensure efficient and reliable data exchange between processes. Because the VLM runs in an independent process, the chat assistant can be safely terminated without causing a crash or leaving GPU memory allocated. Closing the chat window stops communication with the subprocess, which releases all associated resources upon exit.

*Model Management and Performance Optimization.* The worker subprocess is responsible for loading and executing the selected VLMs from the Hugging Face Hub. To make these large-scale models accessible on consumer-grade hardware, OnSight Pathology employs on-the-fly model quantization leveraging the bitsandbytes (v0.45.4)[44] library. The VLM weights are loaded in reduced precision format (e.g., 4-bit or 8-bit), substantially lowering VRAM requirements and enabling operation on GPUs with limited memory capacity. For better interactivity, OnSight Pathology implements real-time token streaming from the worker process to the chat interface. Model responses are transmitted incrementally, token-by-token, as they are generated, rather than waiting for a completed full response. The GUI renders their sequential text chunks immediately, producing a "live typing" effect that creates a more natural conversational experience for the user.

**Benchmarking Performance Across Different Devices.** To evaluate OnSight Pathology's performance and demonstrate its compatibility across a wide range of consumer-grade systems, benchmarking was conducted on personally-owned devices among team members and early adopters of the platform. Users were instructed to select the recommended capture area for each model and record the latency values displayed in the inference window. For each hardware configuration, 10 latency readings were sampled and used to compute the mean and standard deviation of the inference time. The make and model of the CPU or GPU for each device were also documented.

**Comparing Ki-67 Estimates Across Different Tools.** To benchmark OnSight Pathology's performance in computing Ki-67 proliferation indices, a dataset of 17 Ki-67-stained tissues, with available pathologist-derived estimates, was used[45]. For each case, a representative region was selected, and the Ki-67 index was calculated using OnSight Pathology's nuclear segmentation model. For comparison, the same regions were analyzed using two alternative tools: QuPath, a widely used desktop application for digital pathology, and DeepLIIF, an online deep learning-based platform that provides automated Ki-67 quantification. The Pearson correlation coefficient was calculated to assess concordance between the outputs from each tool and the pathologist's estimates (**Fig 3j**). To further illustrate OnSight Pathology's workflow efficiency, we compared the time required to perform each analysis. Unlike other programs that require manual image extraction or uploads to external servers, OnSight Pathology computes the index directly within the viewer. The time required for these additional preparation steps was recorded for all 17 images (**Fig 3k**), and the mean and standard deviation were calculated and compared with the time required to obtain results using OnSight Pathology.

**Validation and Benchmarking Cohorts.** To validate OnSight Pathology's ViT-based tumor classification model for the four common neuropathology tumor histological patterns (glioma, meningioma, schwannoma, metastatic epithelial tumors), we used the TCGA online slide viewer and the EBRAINS resource. "The Digital Brain Tumour Atlas (v1.0)," was chosen for its widespread use and breadth across many subclasses. We tested 201 TCGA-LGG and 201 TCGA-GBM cases, and the epithelial class was formed by combining 200 TCGA-COAD, 220

TCGA-BRCA, and 201 TCGA-LUAD cases (621 epithelial cases total). In EBRAINS, the in-distribution set comprised 1,409 cases aligned to our four categories (943 gliomas across 11 subclasses, 344 meningiomas across 7 subclasses, 75 schwannomas across 2 subclasses, 47 metastatic epithelial tumors), while the out-of-distribution set comprised 211 cases (28 chordomas, 20 germinomas, 13 papillary craniopharyngiomas, 50 adamantinous craniopharyngiomas, 50 hemangioblastomas, 50 diffuse large B-cell lymphomas), yielding 1,620 EBRAINS cases in total and covering more than 50% of the reported Digital Brain Tumor Atlas patient cohort. All slides were reviewed at 20× magnification, representative histology regions were selected for each case, case-level predictions were tallied with the Aggregate function in OnSight Pathology, and performance was summarized using confusion matrices (**Fig 3e**). Multi-Class ROC curves were generated to visualize the performance of the model for each class (**Fig 3d**).

For the external evaluation of our mitosis-detection RetinaNet model, we used the ICPR 2012 and ICPR 2014 Mitosis Grand Challenge datasets. Source images were kept in their native BMP or JPG formats without conversion, and the analysis magnification was set to the value optimized for the OnSight Mitosis Detection RetinaNet model. Each case was scanned thoroughly with OnSight Pathology, and all proposed mitoses, with confidence scores, were recorded. Using the ICPR-supplied annotations as a reference, detections matching annotated mitoses were counted as true positives and non-matching detections as false positives, according to the challenge protocols. In total, we recorded 251 detections for ICPR 2012 and 234 detections for ICPR 2014. Youden's J statistic was used to select the optimal threshold. Precision, recall, and the corresponding F-measure were then calculated for comparison. We plotted ROC curves to visualize the results for ICPR 2012 (**Fig 3g**) ICPR 2014 (**Supplementary Fig 11**) challenges. OnSight Pathology's performance was also summarized in a comparative table alongside top ICPR 2012 (**Fig 3h**) and 2014 (**Supplementary Fig 12**) challenge entries.

**Validation of OnSight Pathology in Local Clinical Cohorts.** In addition to extensive validation of OnSight Pathology using verifiable public datasets, we also evaluated its performance within our local clinical workflow using real-world examples.

*Validation on Local Whole-Slide Images.* We evaluated the ViT-B/16 classifier on 98 local clinical cases viewed within the institutional Synapse Pathology viewer (Fujifilm). Representative ROIs were selected at 20× magnification and analyzed in real time with OnSight Pathology. Tile-level predictions were aggregated to the case level using the built-in Aggregate function. The four tissue categories were: glial, meningothelial, metastatic epithelial, and schwannian histology. Performance summaries (e.g., confusion matrix) were computed at the case level (**Fig 3f**).

Ground truth labels for the surgical pathology diagnosis of each case were generated and finalized following review at our multidisciplinary tumor board (neuropathology, neurosurgery, neuroradiology, neuro-oncology) before any model runs. During inference runs, analysts were reader-blinded to tumor-board diagnoses. Labels were revealed only for scoring. Diagnoses that fell outside the four major tumor types of the ViT model were considered out of distribution cases for the analysis.

Local benchmarking of the mitosis model was carried out using pre-annotated WSI where mitoses were identified for formal grading. The model sensitivity was set to 0.99, and all boxed nuclei were recorded with their confidence scores and classification as either true mitoses or false positives. The ROC curve and corresponding AUC were used to benchmark the performance of OnSight Pathology, with the optimal threshold determined using Youden's statistics (**Fig 3i**).

Local benchmarking of the Ki-67 index was carried out using the aggregator function which averaged scores over representative regions. The generated score was compared to the value reported in the final pathology report, and their correlation was evaluated using Pearson's r and corresponding p values (**Fig 3l**).

**Benchmarking Outputs of Multimodal Vision Language Models.** To explore the potential synergy between VLMs and rigid image classification models, we analyzed representative in-distribution and out-of-distribution cases drawn from the EBRAINS cohort. The in-distribution examples consisted of gliomas, whereas the out-of-distribution examples included diffuse large B-cell lymphomas and cortical tissue. These out-of-distribution cases were chosen because the ViT classifier frequently assigned them extremely high classification scores for glioma, reflecting its limited ability to recognize distinct morphologies outside its training domain. We believe this provided an exciting use case for a complementary vision language model.

For each selected image, the incorporated VLM Lingshu-7B was prompted to provide a microscopic textual description to assess if there were consistent differences in phrasing or terminology between the distinct classes. In addition to including some representative descriptions, we also took a more objective approach by converting the generated descriptions into high-dimensional vector representations using the SentenceTransformer library (v5.1.1, https://github.com/UKPLab/sentence-transformers), with the pretrained *all-MiniLM-L6-v2* model, and visualized their embeddings by t-SNE (**Fig 4i**). This allowed us to observe how samples from the same ground-truth labels clustered relative to one another. Samples of normal cortical tissue from a local case were also included for reference. As a complementary approach, we also computed the pairwise cosine similarity between these text embeddings to create a similarity matrix. This matrix was then visualized as a hierarchical clustering heatmap using Ward's linkage method[46] (**Fig 4j**).

**Testing OnSight Pathology on a Live Camera Feed of Glass Slides.** In addition to classifying images of digitized WSIs viewed through a slide viewing software, we reasoned that the presented workflow should theoretically also work on live images of glass slides projected onto a screen using a camera. To test this, we took advantage of our clinical telepathology interoperative consultation system that uses the Mikroscan digital pathology system. This system projects FOVs from glass slides onto a remote desktop in our pathology department at a different hospital site about 3-5 km from our neurosurgical operating rooms (**Fig 5a**). To assess intra-operative use, we analyzed frozen sections from 9 patients (glial n = 2, meningothelial n = 2, epithelial n = 3, schwannian n = 2). A total of 27 extracted images were viewed at 20x and the autofocus was applied prior to inference generation (**Supplementary Fig 13**). Results were summarized with a confusion matrix (**Fig 5d**).

**Compatibility of OnSight Pathology with Smartphone Camera Live Feed.** We further demonstrate OnSight Pathology's compatibility with live microscope imaging via a smartphone camera. For this demonstration, the smartphone camera was positioned over the microscope eyepiece using the OpenOcular OE2 (OpenOcular, Florida, USA) smartphone-to-microscope adapter (**Fig 5f**). The STL files for the OE2 adapter were downloaded from the OpenOcular website, the G-code was generated using Slic3r Prusa (v1.41.3, https://slic3r.org/) and was then printed out of 1.75 mm polylactic acid (PLA) filament using the Prusa i3 MK3 (Prusa Research, Czech Republic) 3D printer. The print time was 16 hours, and the total material cost was $2.54 CAD. With the camera running, the phone was connected to a Windows 11 laptop via a USB-C cable to a video-enabled USB-C port, and the phone screen was mirrored to the laptop screen using SCRCPY (v3.3.3, https://github.com/Genymobile/scrcpy). This setup enables the view from the microscope eyepiece to be projected to a laptop without the need for expensive specialized equipment. OnSight Pathology was then used to demonstrate real-time automated histologic analysis of the histology slide from a live smartphone camera feed (**Fig 5e**).

**Acknowledgements:**
This work is supported by the Canadian Institutes of Health Research Project Grant (FRN: 178104) (P.D.), a Cancer Research Society Operating Grant (1280470) (P.D.) and the Princess Margaret Cancer Foundation (P.D.).

**Author Contributions:**
J.H., K.F. and P.D. conceived the idea and approach. K.F. and J.H. developed the software and computational pipelines. P.B.Z., A.B., D.G.O. and A.P. analyzed and interpreted the data outputs for the validation studies. S.E., A.Y., A.A., D.G.O., E.R.K and A.S.A. contributed to the design, clinical implementation approach and user interface. J.H., K.F., P.B.Z. and P.D. wrote the manuscript with input from all other authors. P.D. supervised the work.

**Conflict of Interest:**
The authors declare no competing interests.

**Data Availability:**
The code for OnSight Pathology is available on Github and as an executable file at *https://github.com/JinzhenHu/OnSight_Pathology*. As benchmarking was largely done on public datasets and through local WSI viewers without direct access to digital images, no new datasets were generated in this study.

**References:**
1. Louis, D. N. *et al.* The 2021 WHO Classification of Tumors of the Central Nervous System: a summary. *Neuro Oncol* **23**, 1231–1251 (2021).
2. Chen, P. H. C. *et al.* An augmented reality microscope with real-time artificial intelligence integration for cancer diagnosis. *Nat Med* **25**, 1453–1457 (2019).
3. Chen, R. J. *et al.* Towards a general-purpose foundation model for computational pathology. *Nat Med* **30**, 850–862 (2024).
4. Veta, M. *et al.* Predicting breast tumor proliferation from whole-slide images: The TUPAC16 challenge. *Med Image Anal* **54**, 111–121 (2019).
5. Rimm, D. L. &. *et al.* A Prospective, Multi-institutional, Pathologist-Based Assessment of 4 Immunohistochemistry Assays for PD-L1 Expression in Non-Small Cell Lung Cancer. *JAMA Oncol* **3**, 1051–1058 (2017).
6. Masucci, G. V &. *et al.* Validation of biomarkers to predict response to immunotherapy in cancer: Volume I - pre-analytical and analytical validation. *J Immunother Cancer* **4**, 1–25 (2016).

7. Mok, T. S. K. *et al.* Pembrolizumab versus chemotherapy for previously untreated, PD-L1-expressing, locally advanced or metastatic non-small-cell lung cancer (KEYNOTE-042): a randomised, open-label, controlled, phase 3 trial. *The Lancet* **393**, 1819–1830 (2019).
8. Denkert, C. *et al.* Strategies for developing Ki67 as a useful biomarker in breast cancer. *The Breast* **24**, S67–S72 (2015).
9. Dowsett, M. *et al.* Assessment of Ki67 in Breast Cancer: Recommendations from the International Ki67 in Breast Cancer Working Group. *JNCI Journal of the National Cancer Institute* **103**, 1656–1664 (2011).
10. Martin, B. *et al.* Ki-67 expression and patients survival in lung cancer: systematic review of the literature with meta-analysis. *Br J Cancer* **91**, 2018–2025 (2004).
11. Polley, M. Y. C. *et al.* An international Ki67 reproducibility study. *J Natl Cancer Inst* **105**, 1897–1906 (2013).
12. Djuric, U., Zadeh, G., Aldape, K. & Diamandis, P. Precision histology: how deep learning is poised to revitalize histomorphology for personalized cancer care. *NPJ Precis Oncol* **1**, (2017).
13. Bulten, W. *et al.* Artificial intelligence for diagnosis and Gleason grading of prostate cancer: the PANDA challenge. *Nat Med* **28**, 154–163 (2022).
14. Chen, R. J. *et al.* Pan-cancer integrative histology-genomic analysis via multimodal deep learning. *Cancer Cell* **40**, 865-878.e6 (2022).
15. Faust, K. *et al.* PHARAOH: A collaborative crowdsourcing platform for phenotyping and regional analysis of histology. *Nat Commun* **16**, 742 (2025).
16. Kwon, D. How artificial intelligence is transforming pathology. *Nature* **641**, 1342–1344 (2025).
17. Song, A. H. *et al.* Artificial intelligence for digital and computational pathology. *Nature Reviews Bioengineering* **1**, 930–949 (2023).
18. van der Laak, J., Litjens, G. & Ciompi, F. Deep learning in histopathology: the path to the clinic. *Nat Med* **27**, 775–784 (2021).
19. Qureshi, H. A. *et al.* Synergies and Challenges in the Preclinical and Clinical Implementation of Pathology Artificial Intelligence Applications. *Mayo Clinic Proceedings: Digital Health* vol. 1 Preprint at https://doi.org/10.1016/j.mcpdig.2023.08.007 (2023).
20. Boor, P. Deep learning applications in digital pathology. *Nat Rev Nephrol* **20**, 702–703 (2024).
21. Bankhead, P. *et al.* QuPath: Open source software for digital pathology image analysis. *Sci Rep* **7**, 16878 (2017).
22. Ghahremani, P. *et al.* Deep learning-inferred multiplex immunofluorescence for immunohistochemical image quantification. *Nat Mach Intell* **4**, 401–412 (2022).
23. Chen, J. *et al.* HuatuoGPT-Vision, Towards Injecting Medical Visual Knowledge into Multimodal LLMs at Scale. (2024).
24. LASA Team *et al.* Lingshu: A Generalist Foundation Model for Unified Multimodal Medical Understanding and Reasoning. (2025).
25. ai, kaiko. *et al.* Towards Large-Scale Training of Pathology Foundation Models. (2024).
26. Roetzer-Pejrimovsky, T. *et al.* The Digital Brain Tumour Atlas, an open histopathology resource. *Sci Data* **9**, 55 (2022).

27. Gutman, D. A. *et al.* Cancer Digital Slide Archive: an informatics resource to support integrated in silico analysis of TCGA pathology data. *Journal of the American Medical Informatics Association* **20**, 1091–1098 (2013).
28. Lin, T.-Y., Goyal, P., Girshick, R., He, K. & Dollár, P. Focal Loss for Dense Object Detection. (2018).
29. Aubreville, M. *et al.* https://github.com/DeepMicroscopy/MIDOGpp.
30. Aubreville, M. *et al.* A comprehensive multi-domain dataset for mitotic figure detection. *Sci Data* **10**, 484 (2023).
31. Ludovic, R. *et al.* Mitosis detection in breast cancer histological images An ICPR 2012 contest. *J Pathol Inform* **4**, 8 (2013).
32. Roux, L. & Racoceanu, D. Mitos & Atypia 2014: A Grand Challenge for Breast Cancer Histology Image Analysis. in *Proceedings of the 22nd International Conference on Pattern Recognition (ICPR 2014)* (2014).
33. Redmon, J., Divvala, S., Girshick, R. & Farhadi, A. You Only Look Once: Unified, Real-Time Object Detection. (2016).
34. Ultralytics and Glenn Jocher. Ultralytics YOLO. Preprint at (2025).
35. Faust, K. *et al.* Visualizing histopathologic deep learning classification and anomaly detection using nonlinear feature space dimensionality reduction. *BMC Bioinformatics* **19**, 173 (2018).
36. Holub, P. *et al.* Privacy risks of whole-slide image sharing in digital pathology. *Nat Commun* **14**, 2577 (2023).
37. Hanna, M. G. *et al.* Integrating digital pathology into clinical practice. *Modern Pathology* vol. 35 Preprint at https://doi.org/10.1038/s41379-021-00929-0 (2022).
38. Lee, M. K. I. *et al.* Compound computer vision workflow for efficient and automated immunohistochemical analysis of whole slide images. *J Clin Pathol* **76**, 480–485 (2023).
39. Weinstein, J. N. *et al.* The Cancer Genome Atlas Pan-Cancer analysis project. *Nat Genet* **45**, 1113–1120 (2013).
40. Van Rossum, G. and D. F. L. *Python 3 Reference Manual*. vol. 1441412697 (CreateSpace, 2009).
41. Paszke, A. *et al.* PyTorch: An Imperative Style, High-Performance Deep Learning Library. (2019).
42. Wightman, R. PyTorch Image Models. *GitHub repository* Preprint at https://doi.org/10.5281/zenodo.4414861 (2019).
43. Wolf, T. *et al.* Transformers: State-of-the-Art Natural Language Processing. in *Proceedings of the 2020 Conference on Empirical Methods in Natural Language Processing: System Demonstrations* 38–45 (Association for Computational Linguistics, Online, 2020).
44. Dettmers, T., Lewis, M., Belkada, Y. & Zettlemoyer, L. LLM.int8(): 8-bit Matrix Multiplication for Transformers at Scale. (2022).
45. 10.5281/zenodo.17294901. *Zenodo* Preprint at (2025).
46. Murtagh, F. & Legendre, P. Ward's Hierarchical Agglomerative Clustering Method: Which Algorithms Implement Ward's Criterion? *J Classif* **31**, 274–295 (2014).

# Main Figures:

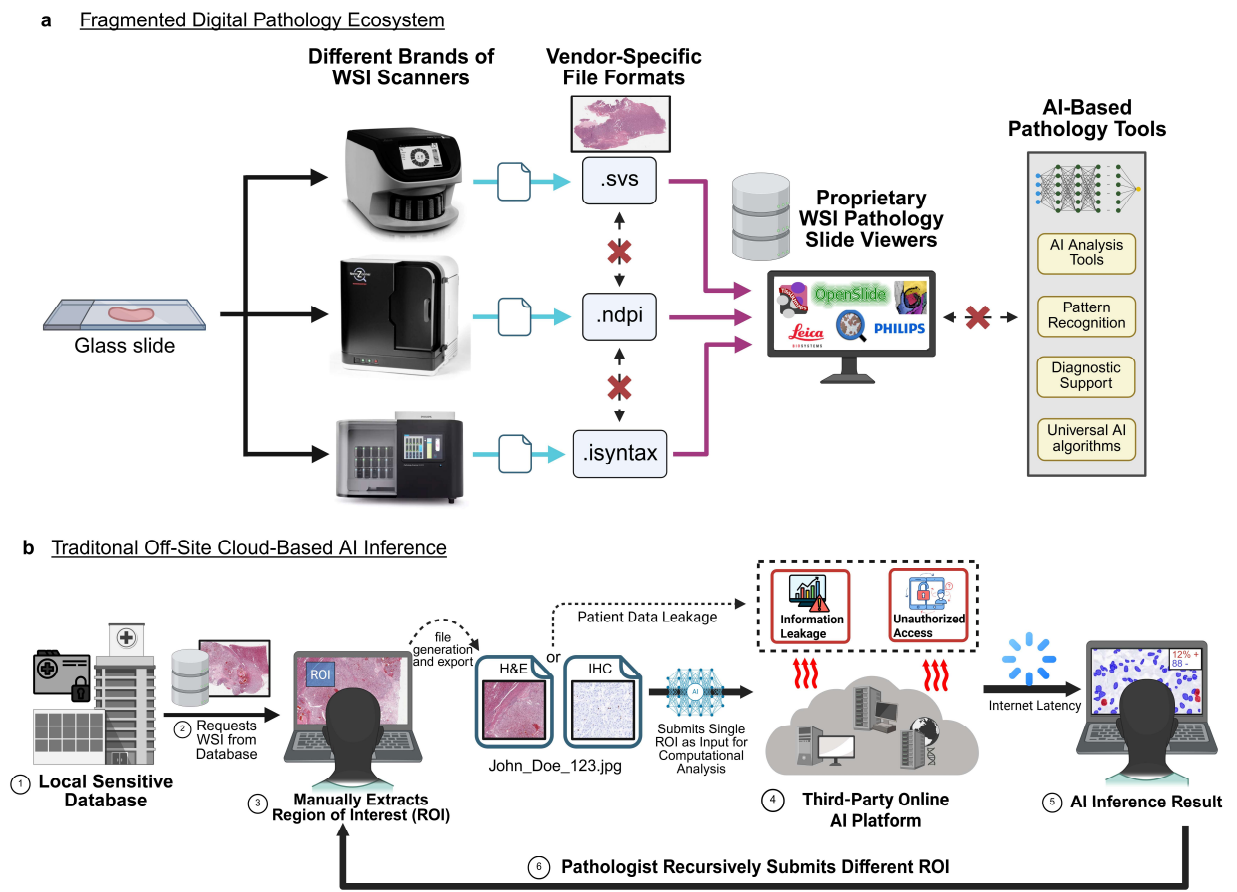

a  Fragmented Digital Pathology Ecosystem

b  Traditonal Off-Site Cloud-Based AI Inference

**Figure 1 | Current challenges and workflow bottlenecks in computational pathology. (a)** The fragmented hardware and software ecosystems in digital pathology challenge adoption. WSI scanners from different manufacturers generate proprietary file formats (e.g., .sys, .ndpi, .isyntax) that are often incompatible with one another and require dedicated slide viewing software. This "vendor lock-in" creates a significant barrier to widespread adoption and deployment of AI algorithms. **(b)** Off-site cloud-based workflows for AI analysis in digital pathology also create workflow inefficiencies and privacy concerns. Pathologists often must manually select, extract, and submit ROIs for computational analysis in a repetitive, multi-step process. This process also risks exposure of patient health information. Internet latency can further delay diagnostic turnaround times and limit clinical practicality.

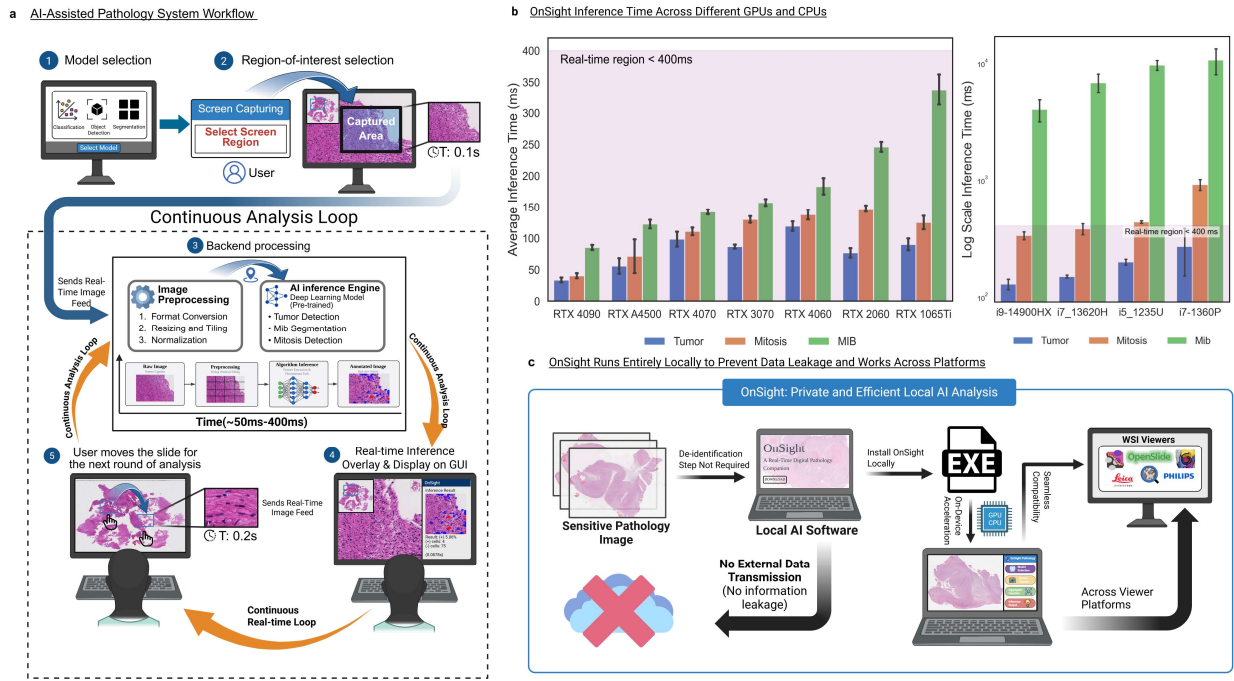

**Figure 2 | OnSight Pathology: A framework for real-time and secure AI in digital pathology. (a)** OnSight Pathology enables real-time, local AI inference directly from the screen capture of images displayed in any standard WSI viewer. The pipeline is initiated when a user selects the desired AI model and uses their mouse to select a custom ROI for continuous screen capture directly from their computer screen. The captured raw image is sent to a local backend, where it is preprocessed and analyzed by the chosen AI model. The inference results are then displayed as an overlay in OnSight Pathology's GUI. To analyze a new area, the user simply moves to their ROI using the WSI viewer, and the entire capture-and-analyze process repeats. This creates a continuous loop, achieving real-time AI analysis of digital pathology images. **(b)** Laptop GPU performance (Left Plot): When running on consumer-grade laptop GPUs, the average inference latencies for all three tasks (tumor classification, mib segmentation, and mitosis detection) fall below the 400ms range. This performance is considered sufficient for real-time interaction (200ms to 400ms)[2]. Laptop CPU performance (Right Plot): When running on CPU-based laptops, OnSight Pathology's tumor classification and most mitosis detection tasks remain within real-time limits. **(c)** A schematic of data handling protocols highlights differences between OnSight Pathology and other off-site computational pipelines. Traditional cloud-based AI platforms require transmitting sensitive pathology images to external servers, introducing risks of data leakage, metadata exposure, and unauthorized access. OnSight Pathology addresses these issues by performing all computations locally with no patient data transmitted outside the institutional software. Furthermore, OnSight Pathology is designed to be agnostic, operating independently of any specific WSI viewer.

# Sample Real-time Inference Outputs

## Tumor Classification
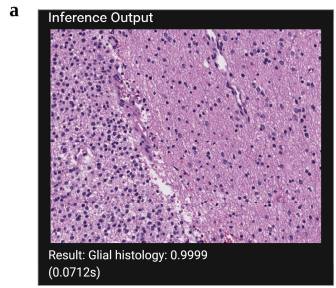

## Mitosis Detection
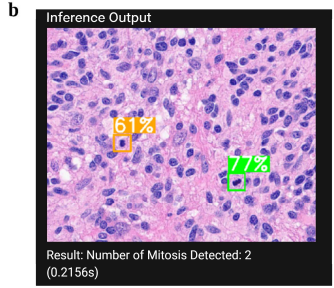

## Mib Segmentation
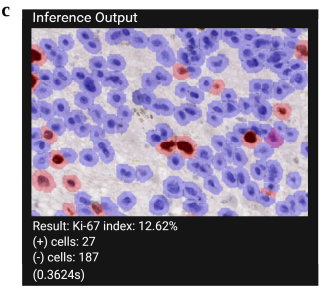

# Benchmarking

## Public Dataset

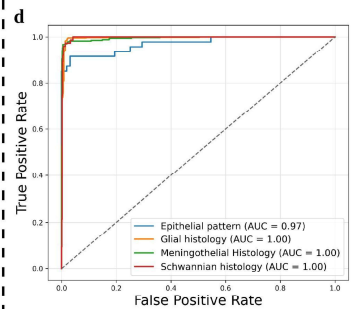

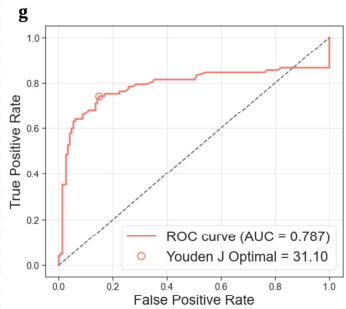

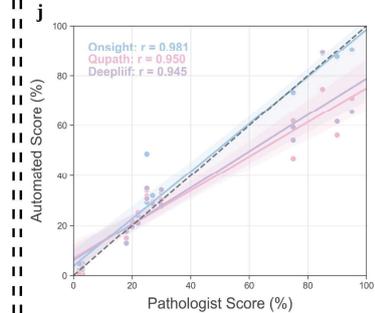

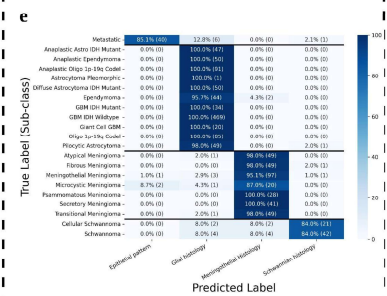

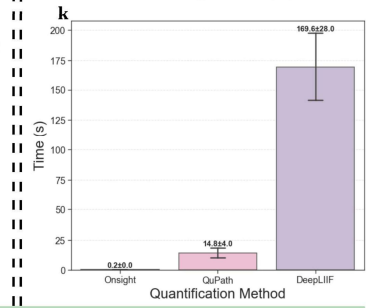

## Local Clinical Workflow

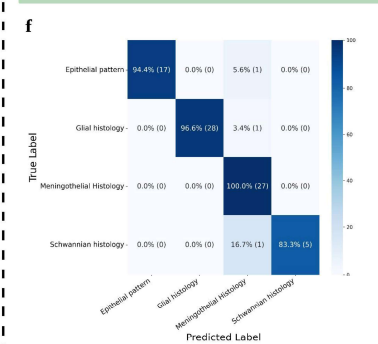

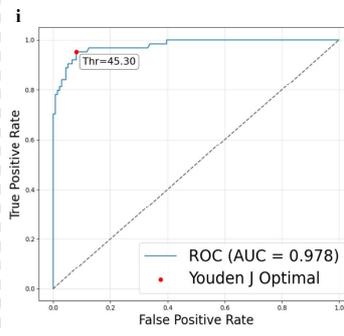

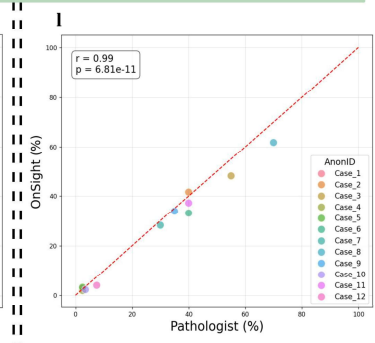

**Figure 3 | Quantitative validation of OnSight Pathology against standard datasets and tools.** Benchmarking for OnSight Pathology was conducted across three tasks: tumor classification, mitosis detection, and MIB (Ki-67) quantification. **(a–**

**c)** Sample OnSight Pathology GUI inference outputs for (a) tumor classification, (b) mitosis detection, and (c) Ki-67 quantification. **(d–e)** ROC curves and confusion matrix for 1,409 "in-class" cases across different common tumor subtypes, **(f)** as well as a confusion matrix for an additional 80 local in-class cases out of a total of 98 evaluated local cases. **(g–h)** Performance metrics on the ICPR 2012 Mitosis Contest dataset, including comparison with others on the contest leaderboard. **(i)** Benchmarking of mitosis detection and confidence scores on local UHN cases. **(j)** Comparison of correlations between OnSight Pathology Ki-67 scores and pathologist-reported counts, versus those obtained using existing tools (e.g., QuPath and DeepLIIF). **(k)** Inference speeds for generating Ki-67 results using OnSight Pathology, compared to other tools that use manually-extracted images as input. **(l)** Correlation between OnSight Pathology-derived Ki-67 estimates and reported indices from pathology reports at our local cancer center (n = 12), with each point representing the mean value across 10–17 sampled regions per case.

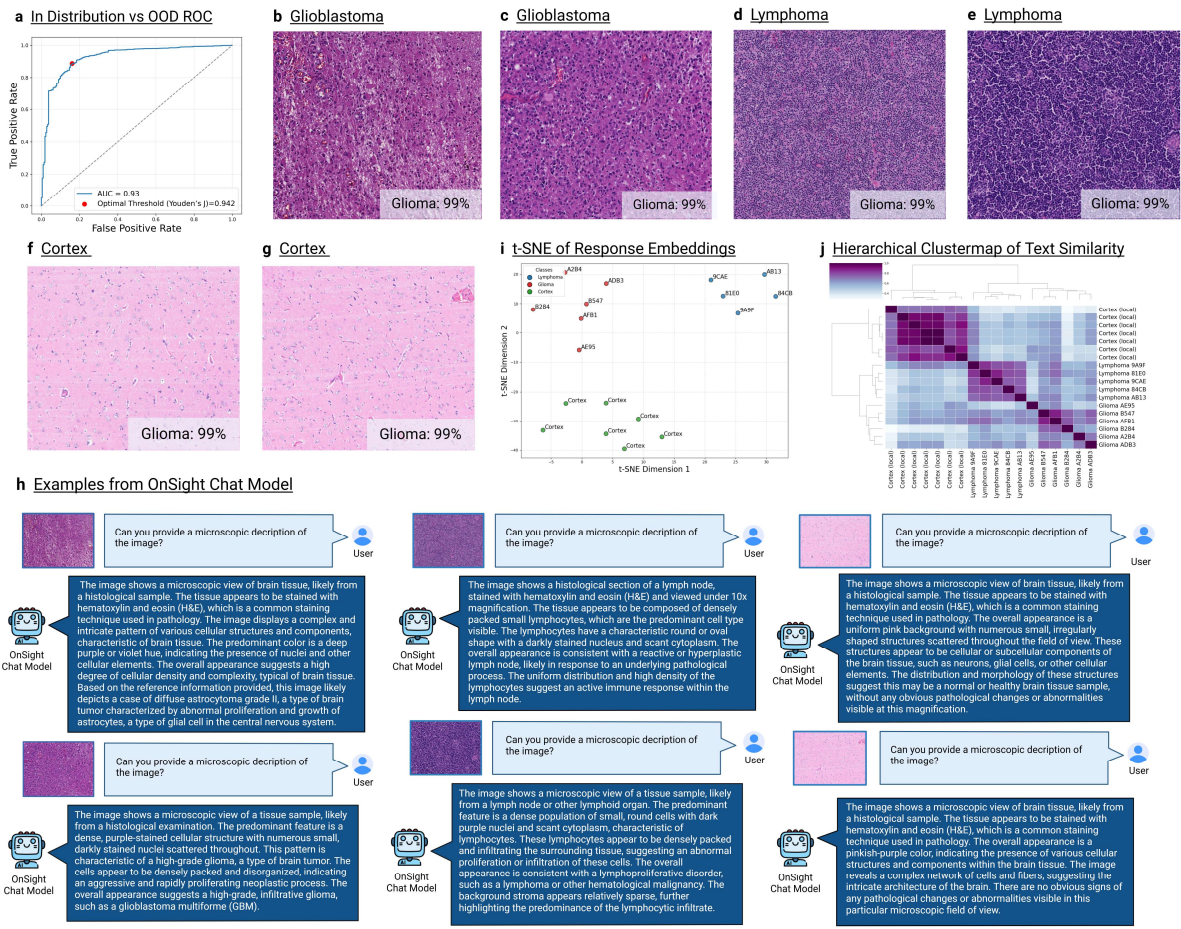

**Figure 4 | Vision language models provide histological descriptions and quality control within OnSight Pathology. (a)** ROC curve for out-of-distribution (OOD) detection using classifier confidence scores. **(b-g)** Representative histology tiles of correctly and misclassified image patches with high confidence scores using the included ViT model **(b-c)** True positive examples where glial histology was correctly classified **(d-g)** Representative OOD (d-e) lymphoma and (f-g) normal cortex image patches incorrectly classified as glioma, each with a confidence score exceeding 99%. **(h)** Sample histological descriptions of images using the integrated VLM. The model generates a unique and relevant histopathological description for each tissue type. For glioblastoma (left), the output references feature of a high-grade tumor; for lymphoma (middle), it correctly identifies lymphocytic morphology of the tissue; and for normal cortex (right), it notes brain tissue with the absence of pathology. **(i)** A t-SNE visualization of sentence embeddings from the chat model's responses to glioblastoma, lymphoma and cortex images. The embeddings form three well-separated clusters, corresponding to each pathology (n=18). This spatial separation confirms that the model's textual outputs are semantically distinct and can be used as an extra quality control to alert the user of OOD tissue types while using OnSight Pathology. **(j)** Hierarchical clustering of the cosine similarity between the textual response embeddings (n=18). The prominent block-diagonal pattern is a result of high

semantic similarity among responses for the same class (intra-class) and low similarity between responses for different classes (inter-class).

**a** OnSight Pathology for Real-Time Telepathology in Frozen Section Diagnosis

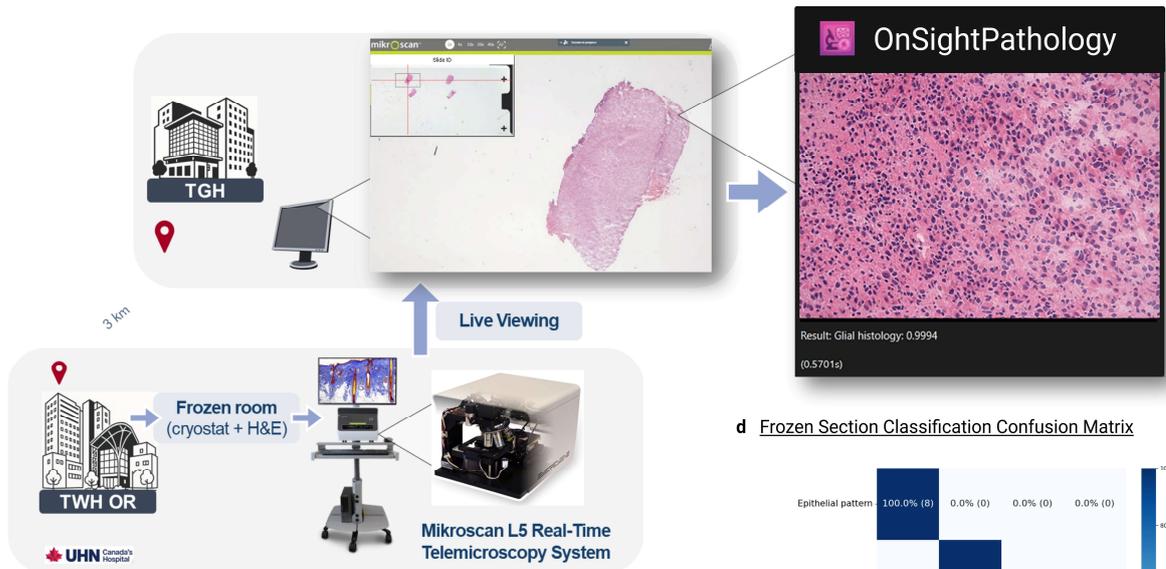

**b** Frozen Section Tumor Classification

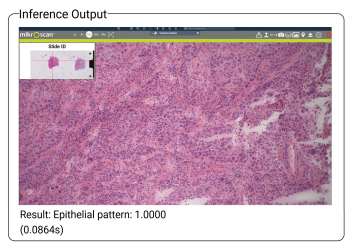

**c** Frozen Section Mitosis Detection

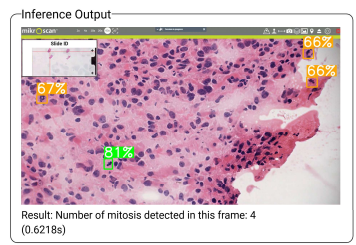

**d** Frozen Section Classification Confusion Matrix

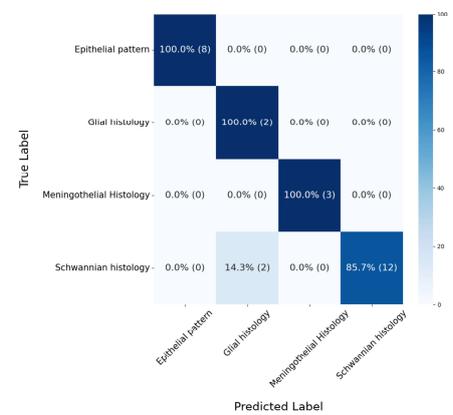

**e** OnSight Pathology Running on Live Microscope Imaging via Smartphone

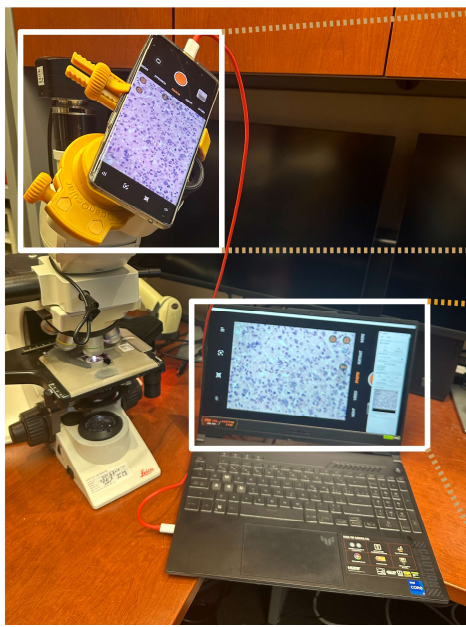

**f** OpenOcular OE2 smartphone-to-microscope Adapter

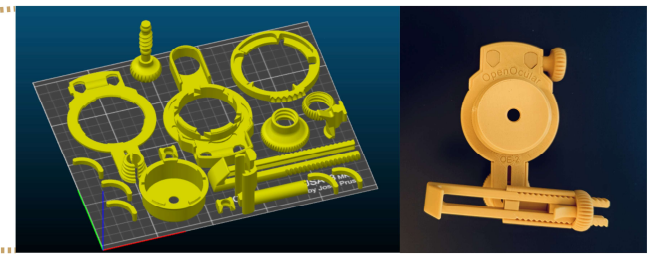

**g** Smart Phone Tumor Detection

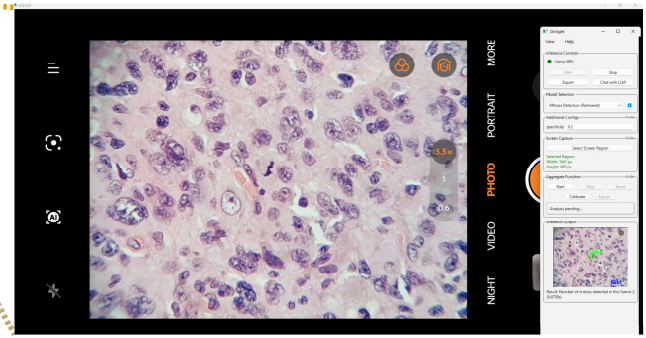

**Figure 5 | OnSight Pathology shows compatibility with live image camera feeds of glass slides imaged with remote telepathology camera systems and smartphone-mounted microscopes. (a)** Diagram of our remote intra-operative telepathology workflow established for frozen section analysis. Glass slides are

prepared next to the operating room at Toronto Western Hospital (TWH) and are streamed remotely from Toronto General Hospital (TGH), about 3 km away, using the Mikroscan real-time telepathology microscopy system. Integration of OnSight Pathology offers additional applications and compatibility with non-WSI setups. **(b-c)** Sample real-time outputs show OnSight Pathology performance on prospective frozen section cases at our cancer center. **(d)** Confusion matrix summarizing OnSight Pathology's classification performance across 27 regions selected from 9 intra-operative patient's frozen section ($n$ = 9).  **(e)** Microscope outfitted with an eye-piece camera using the OpenOcular smartphone mounting system. OnSight Pathology showed combability for tumor classification and mitosis detection with this setup further generalizing the platform to additional potential centers and users. (**f**) The OpenOcular OE2 adapter is shown as a complete 3D print plate (left) and from the front (right). (**g**) Sample of correctly identified mitosis within this distinct setup.

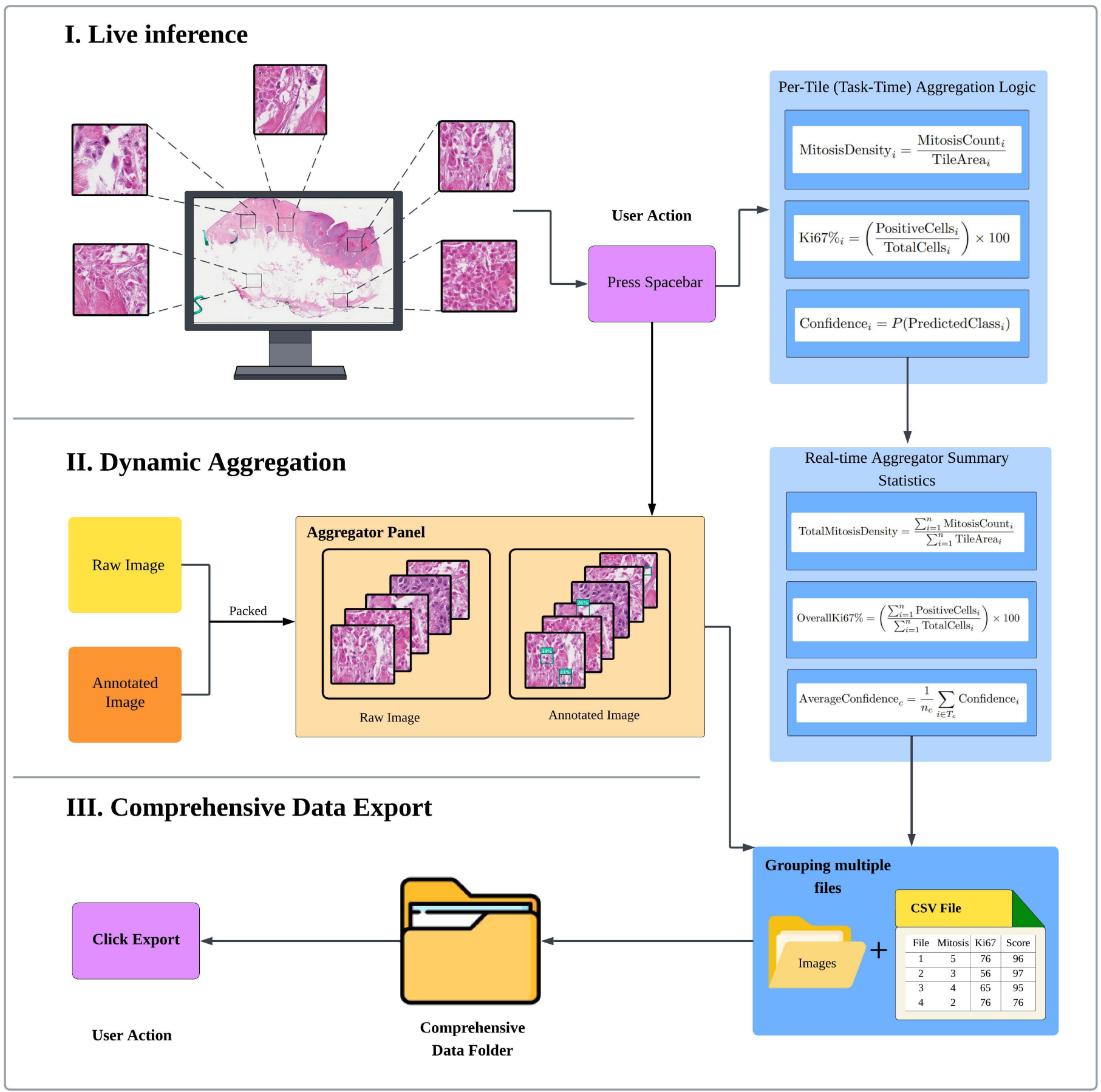

**Supplementary Figure 1 | Aggregating AI Inferences Over Multiple Fields of View.** The aggregator function workflow contains live inference, dynamic aggregation and comprehensive export. **(I)** The live inference process begins with the pathologist identifying and capturing a region of interest directly from the screen using the OnSight Pathology interface. **(II)** With a keypress (spacebar), the captured tile, including both the original raw image and its corresponding AI-inference annotated images, are added to a running list. The aggregator panel updates in real time and the summary metrics are recalculated with each new tile added. **(III)** After analyzing all desired tiles, the user finalizes the session by clicking the "Export" button. This action compiles all captured images and a detailed CSV log file of the session's metrics into a single, organized folder for documentation and further analysis.

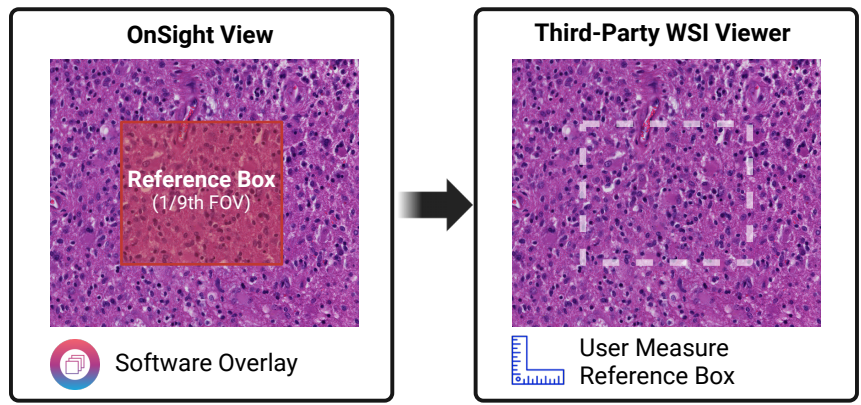

**Supplementary Figure 2 | Physical Unit Calibration in OnSight Pathology.** This two-stage process obtains a precise ROI area by first measuring a reference box using a third-party viewer, then inputting that area for system calculation to determine the total ROI in physical units (mm²).

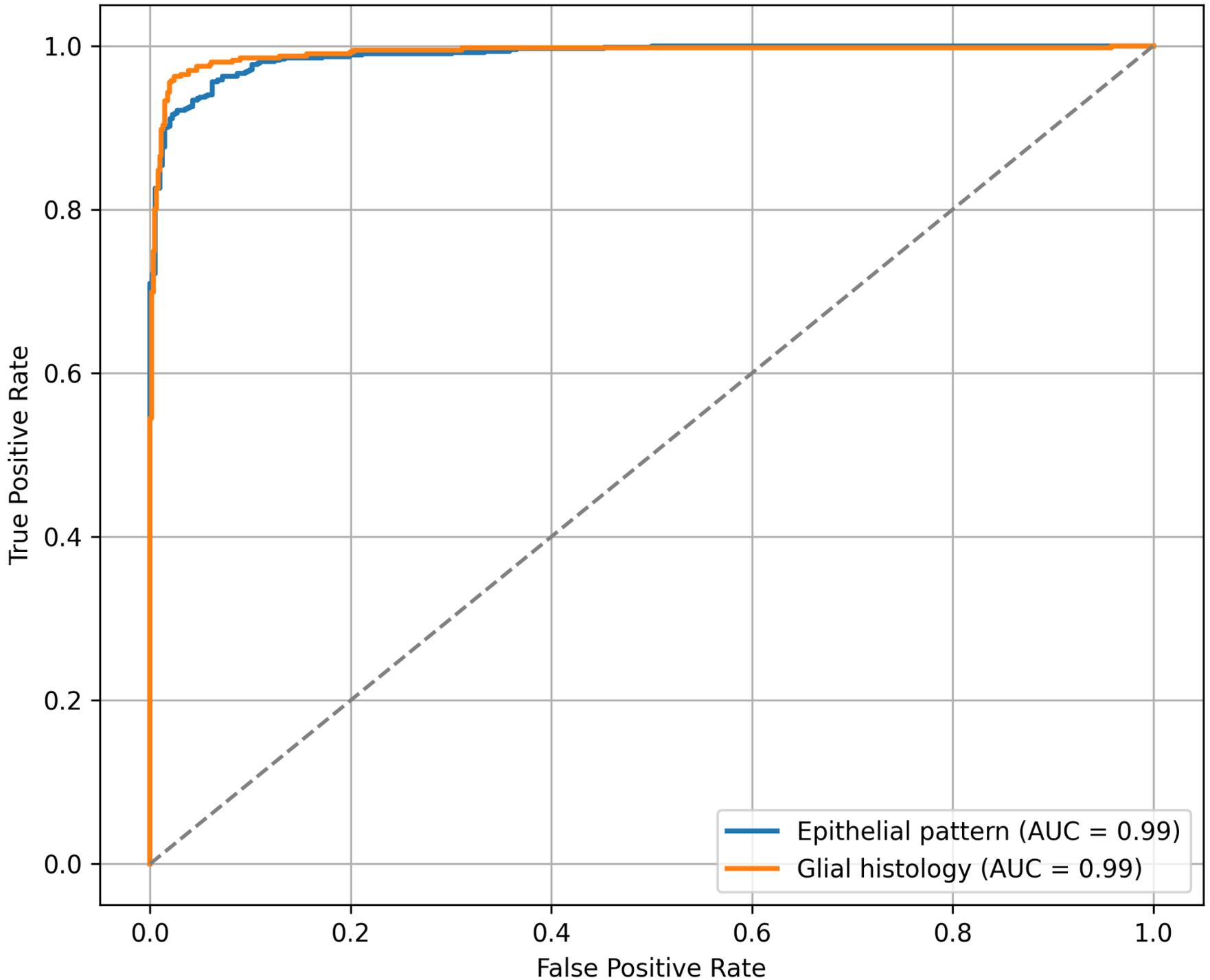

**Supplementary Figure 3 | OnSight Pathology performance across TCGA cohorts.** ROC curves showing OnSight Pathology performance on 1,001 TCGA cases, including glioblastoma (TCGA-GBM), low-grade glioma (TCGA-LGG), and metastatic epithelial tumors, with a multi-class AUC of 0.99.

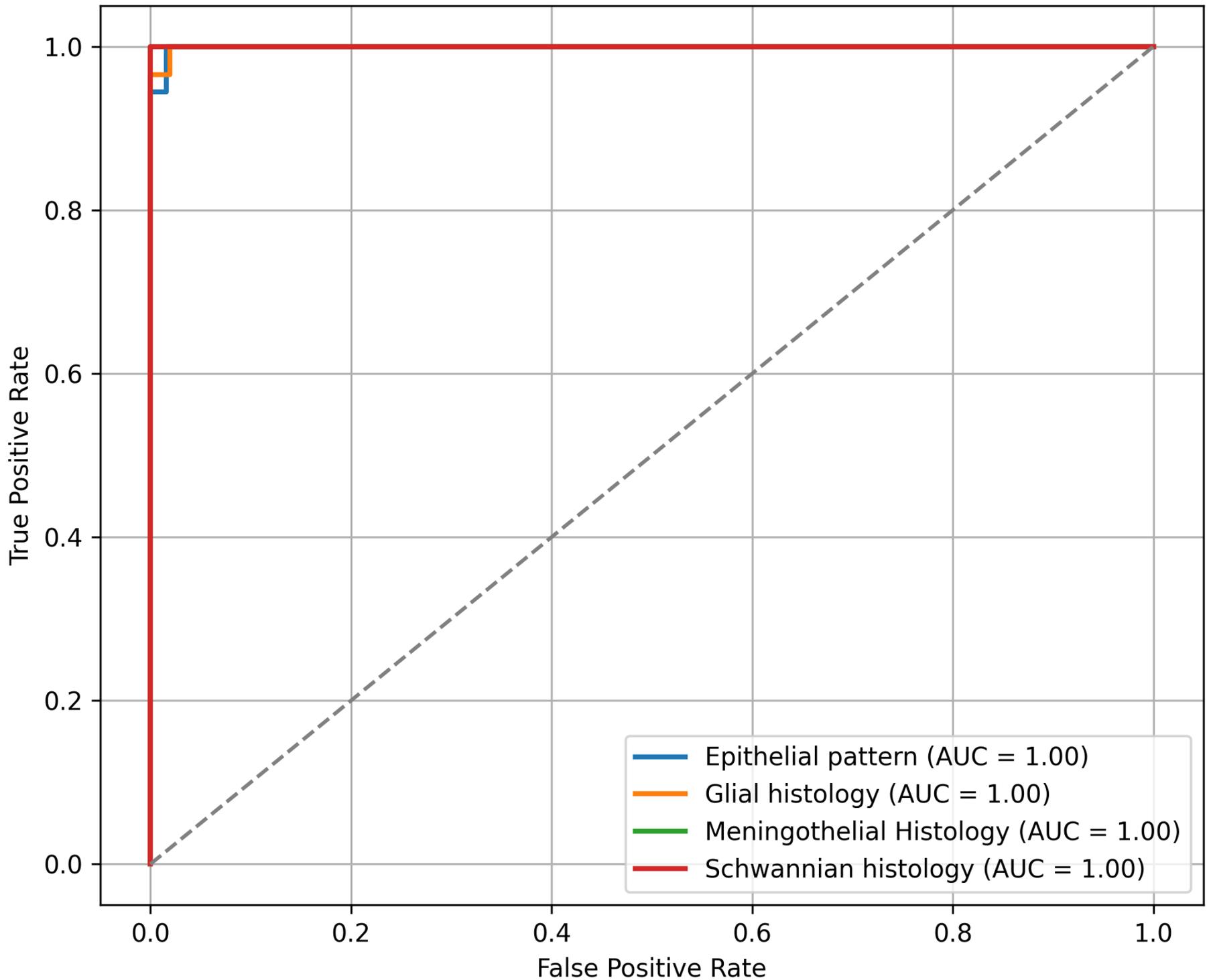

**Supplementary 4 | OnSight Pathology performance across local tumor cases.** OnSight Pathology's performance on 80 in-distribution retrospective tumor board cases (out of 98 total) from the UHN Cancer Center.

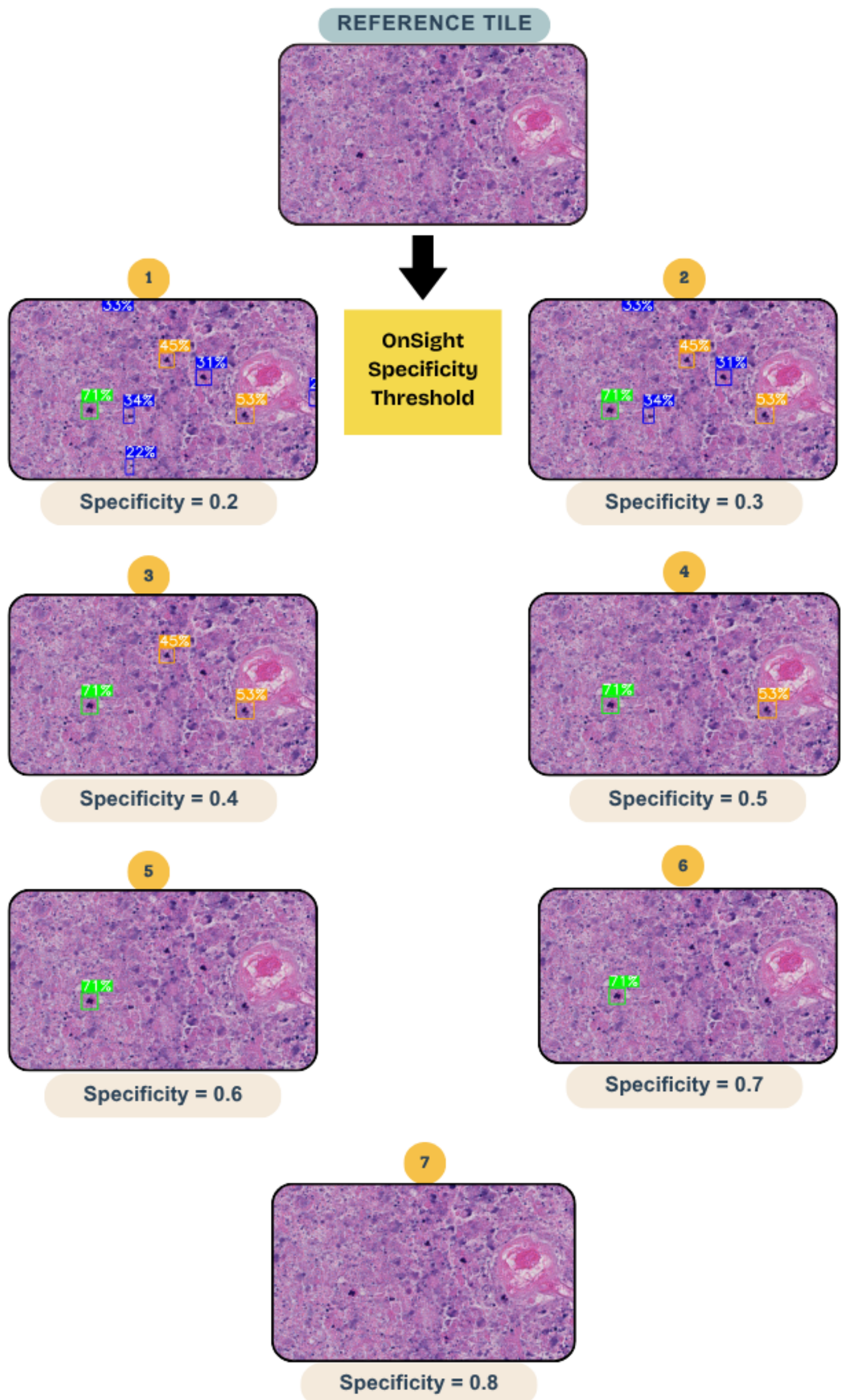

**Supplementary Figure 5 | Real-Time Adjustment of Mitosis Detection Parameters in OnSight Pathology.**
OnSight Pathology features a user-adjustable confidence threshold that modifies the output of the mitosis detection model in real-time. This allows the pathologist to customize the model's sensitivity and specificity based on the specific clinical scenario. For instance, setting a higher confidence threshold makes the model more conservative, displaying only the most certain mitotic figures (higher specificity). Conversely, setting a lower threshold increases the number of candidates detected, maximizing sensitivity.

|  | Example 1 | Example 2 | Example 3 |
|---|---|---|---|
| *Pathologist Alone* | 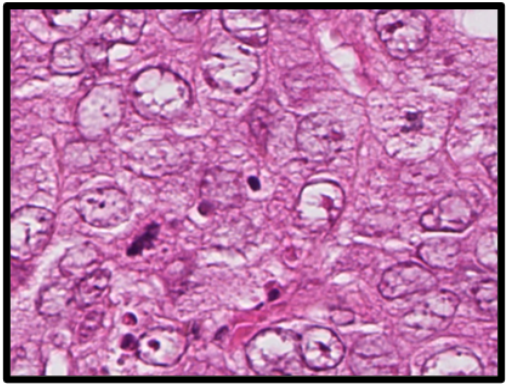 | 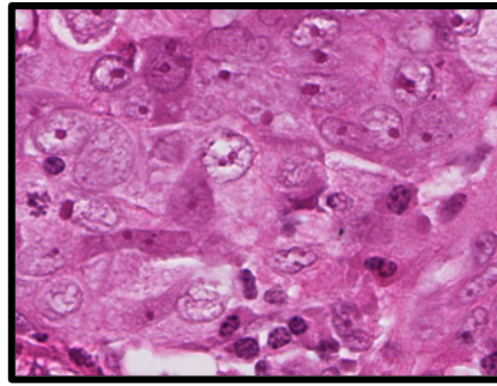 | 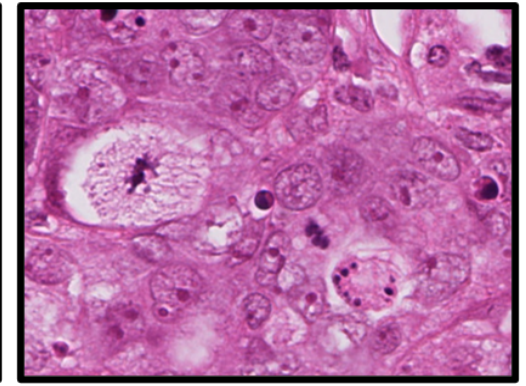 |
| *Pathologist with OnSight* | 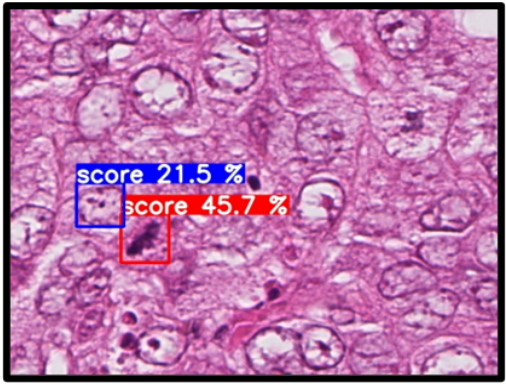 | 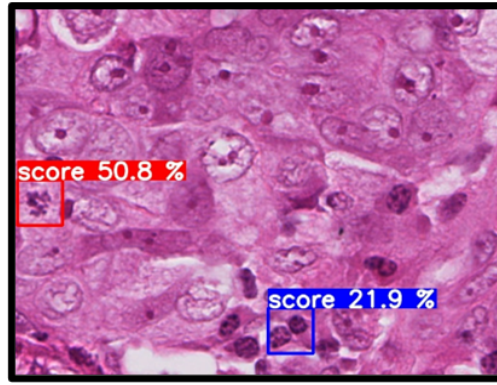 | 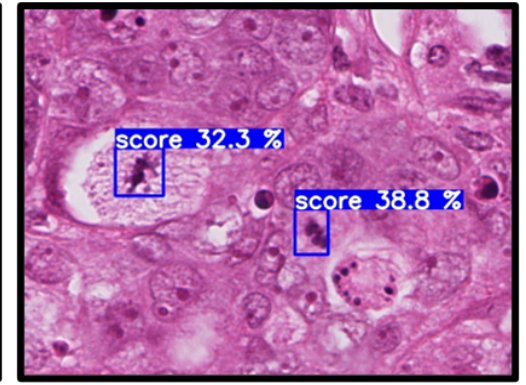 |

**Supplementary Figure 6 | OnSight Pathology Identifies False Negative Mitoses in Public Datasets.** This illustrates the utility of OnSight Pathology in augmenting pathology analysis by identifying true mitotic figures that were missed in the original ground truth dataset (i.e., false negatives). The top row displays three example image patches as they appeared in the original dataset, without any annotations for mitosis. In contrast, the bottom row shows the same patches analyzed by OnSight Pathology, which successfully detected multiple mitotic figures, highlighting them with bounding boxes and confidence scores. These detections represent true mitoses that were previously unannotated.

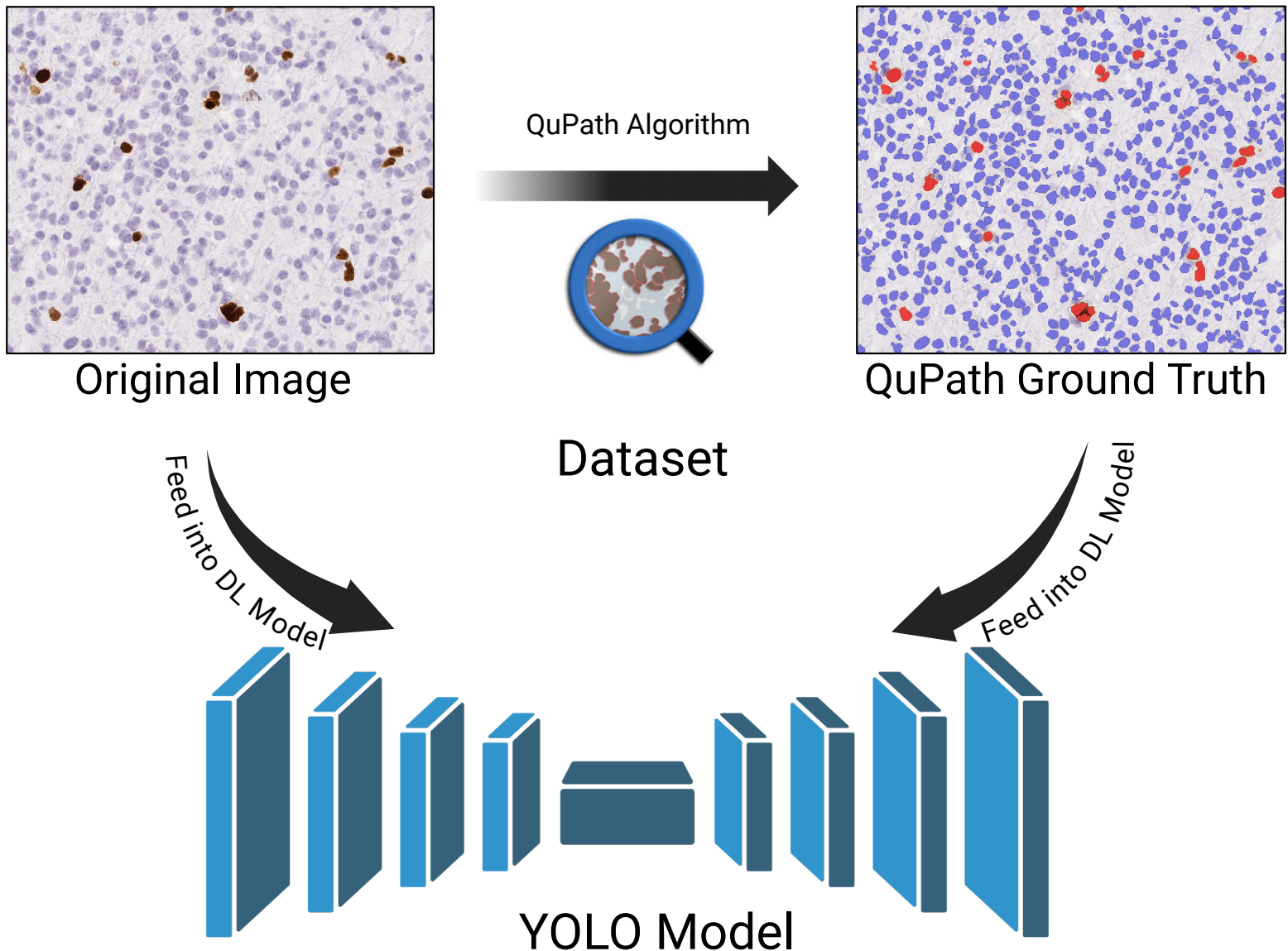

**Supplementary Figure 7 | Ground Truth from QuPath for Ki-67 Quantification Task.** Workflow illustrating 1024x1024 pixel image patches (20x magnification, ~0.5μm/pixel) and their corresponding ground truth annotations, generated using QuPath. These annotated patches serve as training data for a YOLO model for the Ki-67 quantification task.

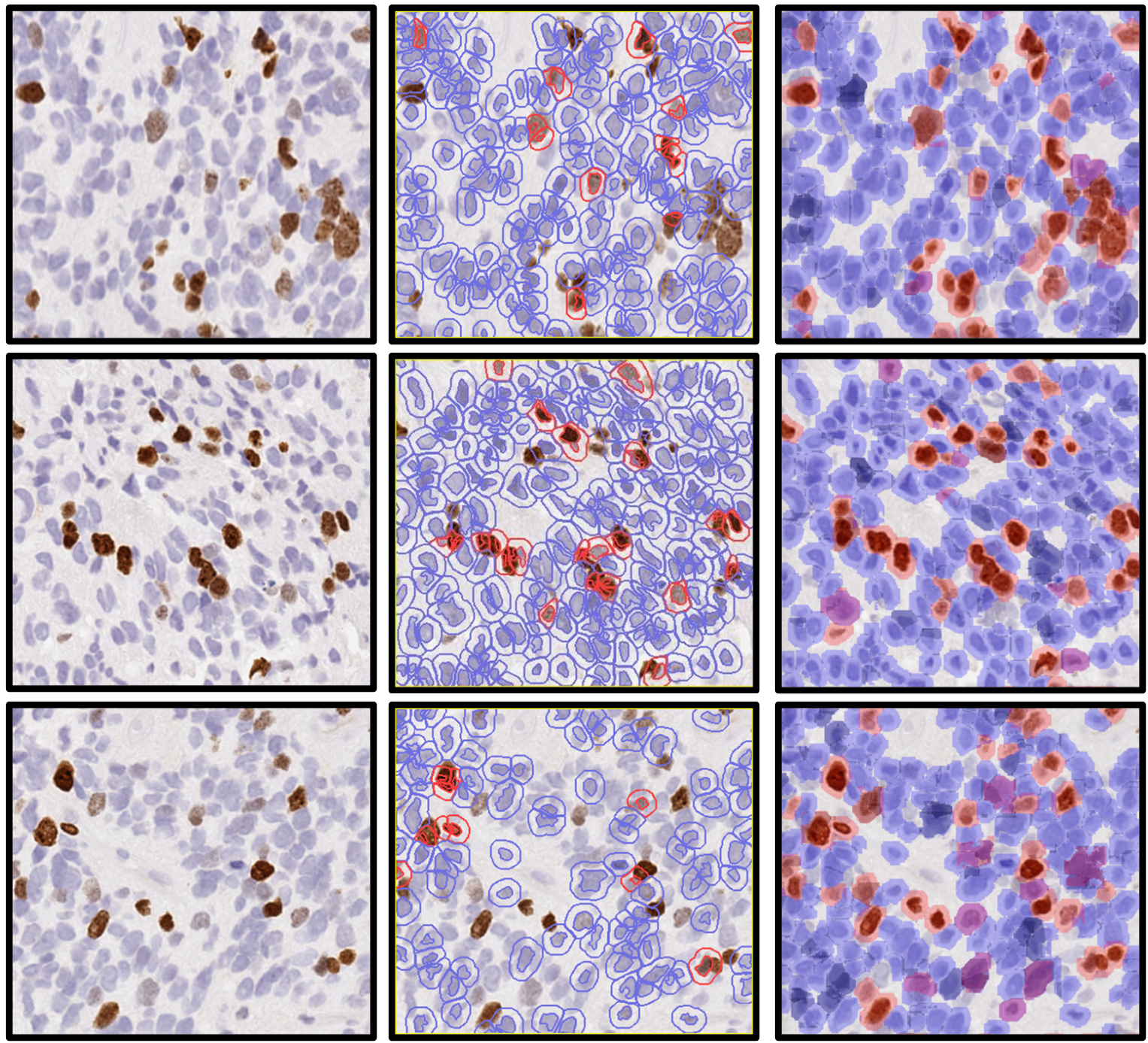

**Supplementary Figure 8 | Challenging Ki-67 Quantification Cases.** OnSight Pathology demonstrates high consistency and accuracy even in challenging Ki-67 quantification cases.

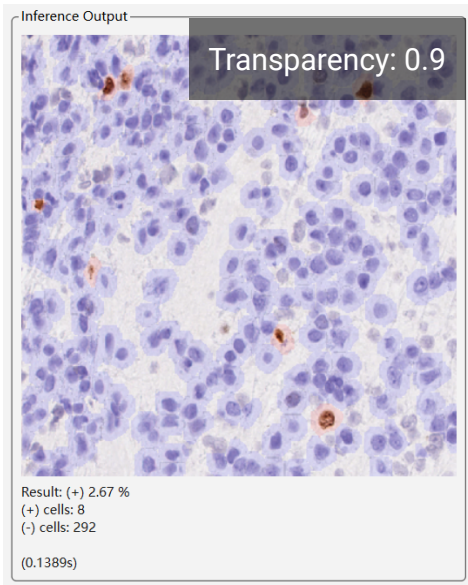 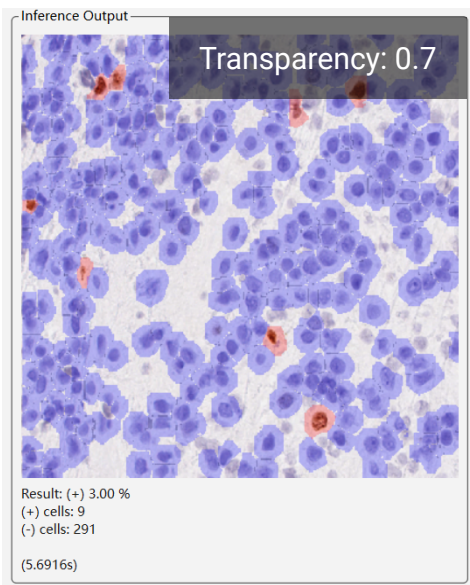 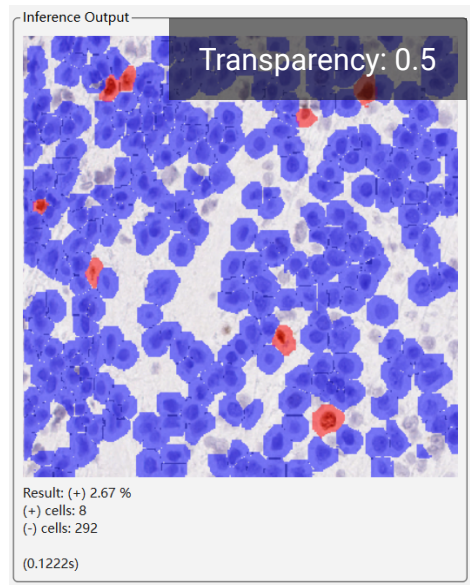 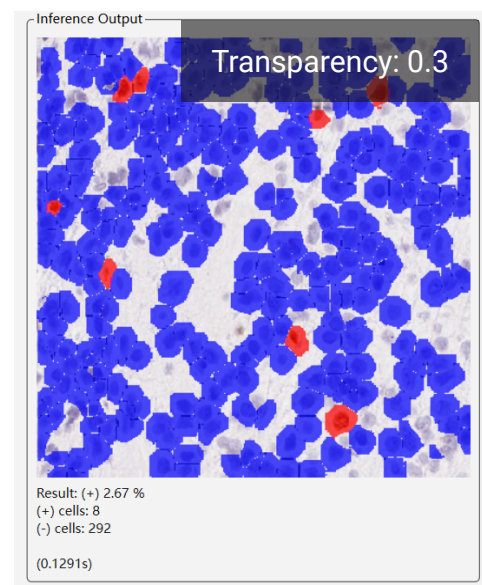

**Supplementary Figure 9 | Interactive Transparency Control for Overlaid Segmentation Visualization.** OnSight Pathology includes a real-time transparency adjustment tool that enables users to modulate the opacity of overlaid segmentation masks for improved interpretability. By adjusting the transparency parameter (0.3, 0.5, 0.7, 0.9), users can dynamically balance visibility between the segmentation overlay and the underlying cellular morphology, providing clearer examination of nuclei and tissue structures.

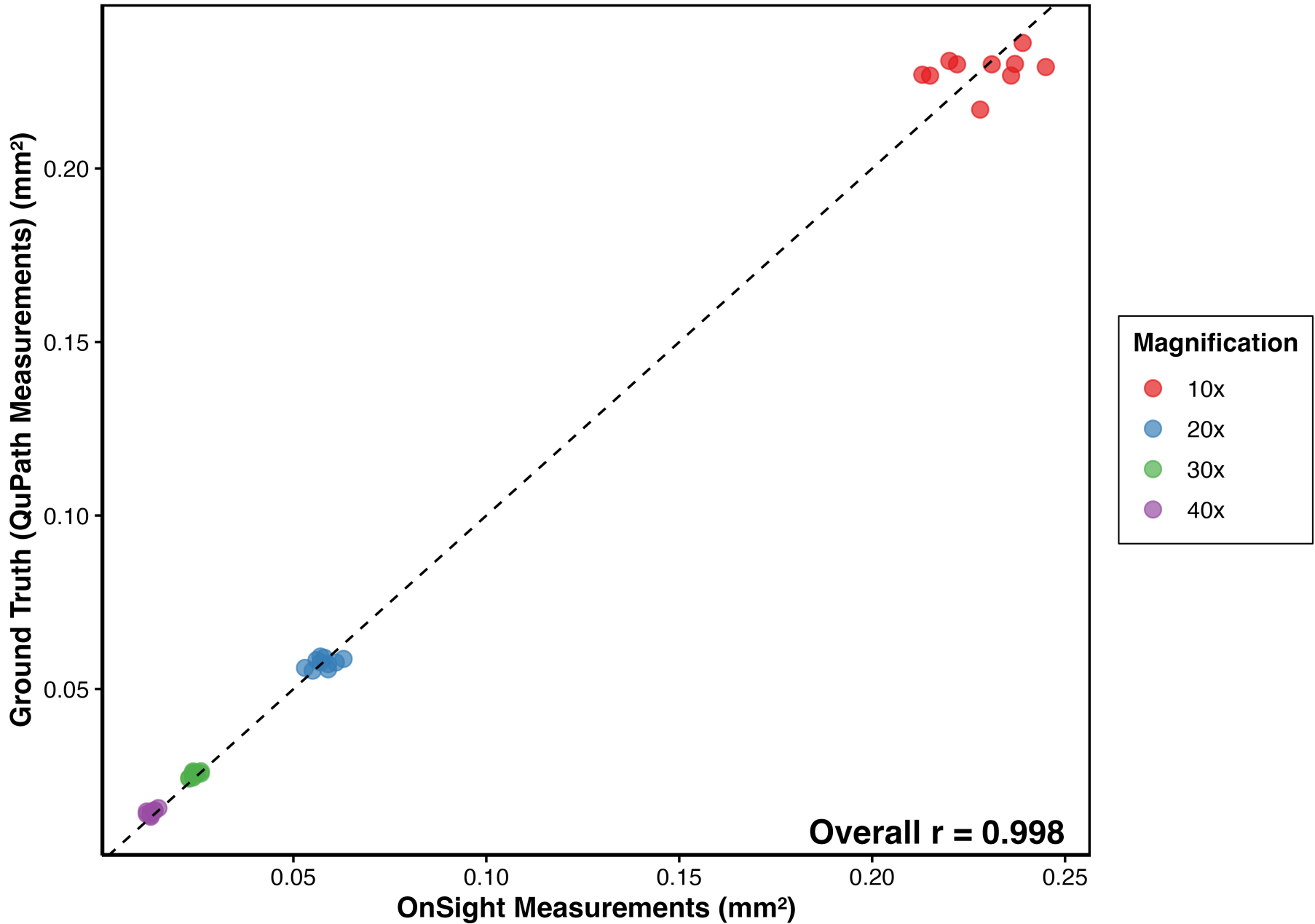

**Supplementary Figure 10 | Benchmarking OnSight Pathology's calibration function.** Correlation plot assessing the reliability of OnSight Pathology's area calibration tool against the ground truth under different magnifications.

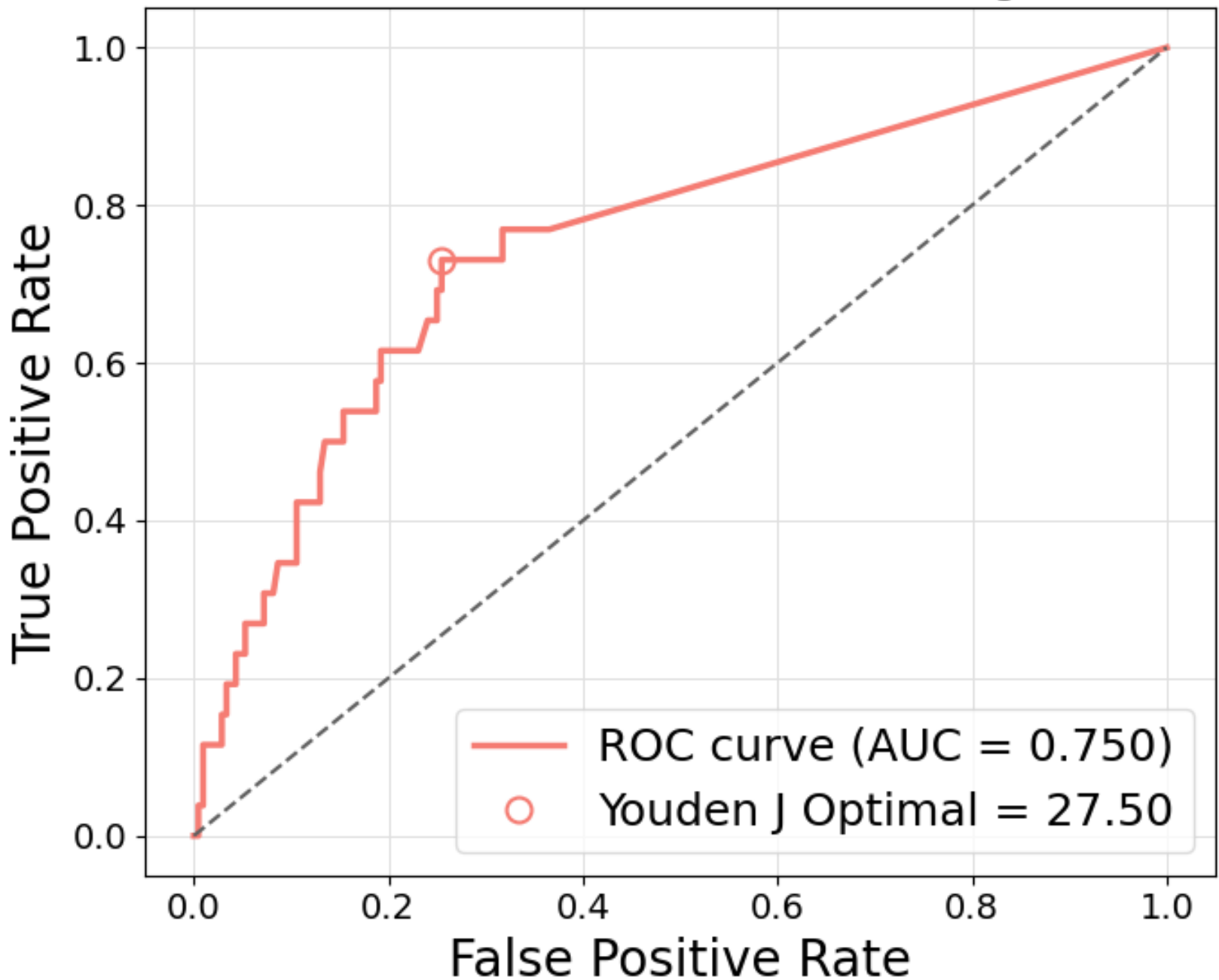

**Supplementary Figure 11 | OnSight Pathology performance on ICPR 2014 mitosis detection challenge** ROC curve for OnSight Pathology's on the ICPR 2014 Mitosis Detection Challenge. The curve shows model performance across confidence thresholds, with the optimal operating point determined using Youden's J statistic (AUC = 0.75).

| Method | F1-score |
|---|---|
| **OnSight** | **0.388** |
| CUHK | 0.333 |
| YILDIZ | 0.172 |
| MINES–CURIE–INSERM | 0.171 |
| STRASBOURG | 0.024 |

**Supplementary Figure 12 | ICPR 2014 Mitosis Challenge Leaderboard.** Summary of F1-scores for the ICPR 2014 Mitosis Detection Challenge. OnSight Pathology achieved the highest performance among the list.

**a** <u>Frozen Section - Metastatic</u>

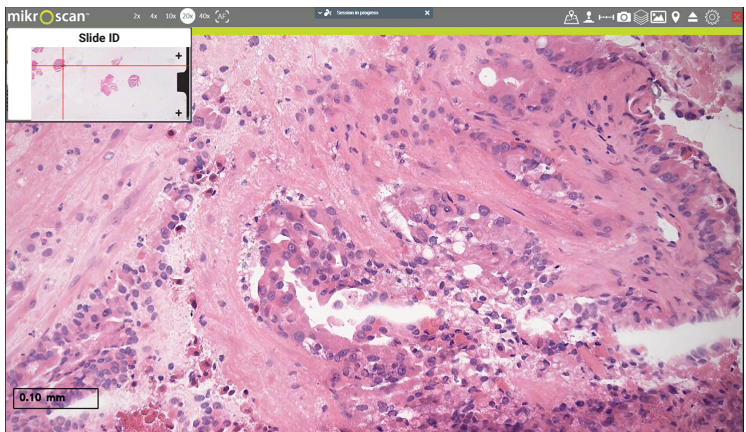

Result: Epithelial pattern: 0.9998

(0.1399s)

**b** <u>Frozen Section- Schwannoma</u>

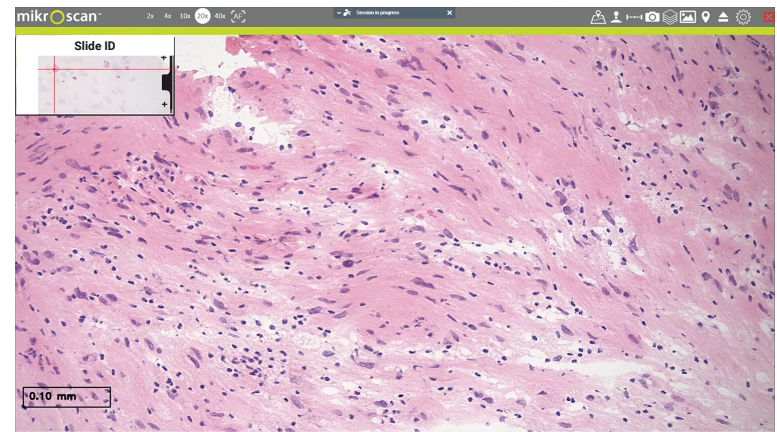

Result: Schwannian histology: 0.9997

(0.1186s)

**c** <u>Frozen Section - Meningioma</u>

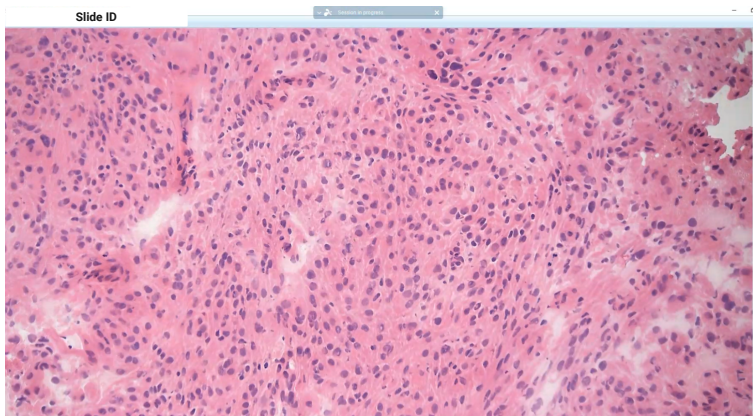

Result: Meningothelial Histology: 0.9938

(0.0706s)

**d** <u>Frozen Section - Glioblastoma</u>

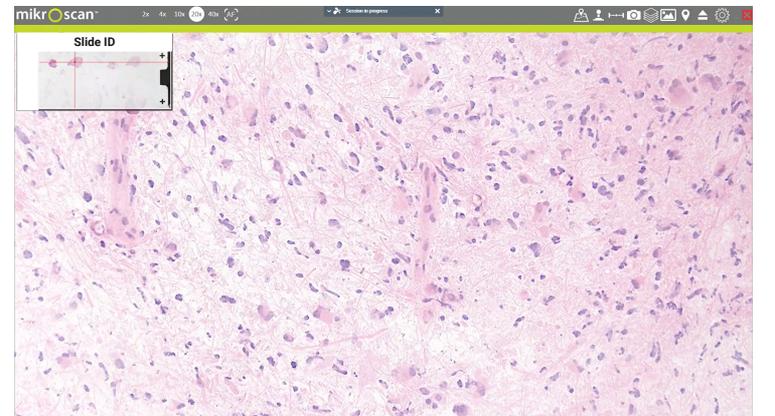

Result: Glial histology: 0.9970

(0.1101s)

**Supplementary Figure 13 | More Real-time Intraoperative frozen section Classification Cases Using OnSight Pathology.** Representative frozen-section demonstrating OnSight Pathology's classification performance on them. Despite the differences in tissue preparation and hardware setup (live camera feeds of glass slides), OnSight Pathology correctly identifies **(a)** an epithelial pattern in metastatic tumor (probability = 0.9998), **(b)** schwannoma (0.9997), **(c)** meningioma (0.9938), and **(d)** glioblastoma (0.9970).